\documentclass{article}

\usepackage[preprint, nonatbib]{main}

\usepackage[utf8]{inputenc} 
\usepackage[T1]{fontenc}    
\usepackage{hyperref}       
\usepackage{url}            
\usepackage{booktabs}       
\usepackage{amsfonts}       
\usepackage{amsmath}
\usepackage{nicefrac}       
\usepackage{microtype}      
\usepackage{xcolor}         
\usepackage{csvsimple}
\usepackage{multirow}
\usepackage{rotating}
\usepackage[square,sort,comma,numbers]{natbib}
\usepackage{graphicx}
\usepackage{subcaption}
\usepackage{lipsum}
\usepackage{wrapfig}
\usepackage{algorithm}
\usepackage{algorithmic}
\usepackage{adjustbox}
\usepackage{rotating}

\newcommand{\bx}{\mathbf{x}}
\newcommand{\by}{\mathbf{y}}

\title{T2FNorm: Extremely Simple Scaled Train-time Feature Normalization for OOD Detection}

\author{
  Sudarshan Regmi$^{1,2,*}$,
  Bibek Panthi$^{1}$, Sakar Dotel, \\ 
  \textbf{Prashnna K. Gyawali$^{3}$, Danail Stoyanov$^{2}$,  Binod Bhattarai$^{4}$} \\ \\
  $^{1}$NepAl Applied Mathematics and Informatics Institute (NAAMII), Nepal\\
  $^{2}$University College London, UK\\
  $^{3}$West Virginia University, USA\\
  $^{4}$University of Aberdeen, UK \\  
  \phantom{\thanks{Corresponding author. Contact: \texttt{sudarshan.regmi@ucl.ac.uk}}}
}
\begin{document}

\maketitle

\begin{abstract}
    Neural networks are notorious for being overconfident predictors, posing a significant challenge to their safe deployment in real-world applications. While feature normalization has garnered considerable attention within the deep learning literature, current train-time regularization methods for Out-of-Distribution(OOD) detection are yet to fully exploit this potential. Indeed, the naive incorporation of feature normalization within neural networks does not guarantee substantial improvement in OOD detection performance. In this work, we introduce \textbf{T2FNorm}, a novel approach to transforming features to hyperspherical space during training, while employing non-transformed space for OOD-scoring purposes. This method yields a surprising enhancement in OOD detection capabilities without compromising model accuracy in in-distribution(ID). Our investigation demonstrates that the proposed technique substantially diminishes the norm of the features of all samples, more so in the case of out-of-distribution samples, thereby addressing the prevalent concern of overconfidence in neural networks. The proposed method also significantly improves various post-hoc OOD detection methods.
\end{abstract}

\section{Introduction}
 The efficacy of deep learning models is contingent upon the consistency between training and testing data distributions; however, the practical application of this requirement presents challenges when deploying models in real-world scenarios, as they are inevitably exposed to OOD samples. Consequently, a model's ability to articulate its limitations and uncertainties becomes a critical aspect of its performance. While certain robust methodologies exist that endeavor to achieve generalizability despite domain shifts, these approaches do not always guarantee satisfactory performance.

OOD detection approaches can be broadly grouped into three approaches: post-hoc methods, outlier exposure, and training time regularization. Post-hoc methods, deriving OOD likelihood from pre-trained models, have significantly improved while outlier exposure, despite the challenges in predefining OOD samples ideally, is prevalently adopted in industrial contexts. Another approach involves training time regularization. This line of work due to its capacity to directly impose favorable constraints during training, potentially offers the most promising path to superior performance. The training-time regularization method, LogitNorm\cite{wei2022mitigating}, employs L2 normalization at the logit level to mitigate overconfidence, leading to an increased ratio of ID norm to OOD norm compared to the results from simple cross-entropy baseline or Logit Penalty\cite{wei2022mitigating}. Nonetheless, the importance of feature norm in achieving ID/OOD separability has been underscored in recent OOD detection works \cite{sun2022out,huang2021importance,tack2020csi,wei2022mitigating}. Normalization at the logit level does not assure an optimal resolution to the overconfidence issue at the feature level. Given the significance of the feature norm, this naturally gives rise to the subsequent query:

\begin{center}
\textit{Can hyperspherical normalization in the higher dimensional feature space provide enhanced benefits without introducing any potential drawbacks?}
\end{center}

Towards this, in this work, we propose feature normalization for OOD detection. Our proposed method requires only a trivial modification to the standard architecture. The feature representation needs to be normalized and scaled during the usual training and inference time. However, the normalization process is intentionally omitted during OOD detection. 
We demonstrate that with the proposed feature normalization, we achieve a clear distinction between the norm of ID and OOD data samples, eventually contributing toward a substantial performance improvement without compromising the model's accuracy.
We show a boost in OOD detection in a number of OOD benchmark datasets (Table \ref{tab:main_table}). For instance, our method reduces the FPR@95 score by \textbf{34\%} with respect to baseline and by \textbf{7\%} with respect to LogitNorm on average across a variety of 9 OOD datasets with DICE scoring on ResNet-18 architecture. In addition, our methods work well in conjunction with many post hoc methods. Our key results and contribution are:

\begin{itemize}
    \item We propose T2FNorm -- a surprisingly trivial yet powerful plug to regularize the model for OOD detection. We quantitatively show that train time normalization approximately projects the features of ID samples to the surface of a hypersphere differentiating it from OOD samples thereby achieving significantly higher \textbf{\textit{separability ratio}}.
    \item We show T2FNorm is equally effective across multiple deep learning architectures and multiple datasets. It also works well in conjunction with multiple post-hoc methods. 
    \item We perform both qualitative and quantitative analysis showing our method's ability to reduce overconfidence and also perform a sensitivity study to show the robustness of our model to the temperature parameter $\tau$. 
    \item We show that \textbf{skipping normalization during OOD scoring time} is a key contributor to our method thus paving the way for exploring the effectiveness of other forms of normalization discrepancies during OOD scoring.
\end{itemize}
\section{Method}

\subsection{Preliminaries: Out of Distribution Detection}
\noindent\textbf{Setup}
Let $\mathcal{X}$ be input space, $\mathcal{Y}$ be output space and $\mathcal{P_{XY}}$ be a distribution over $\mathcal{X} \times \mathcal{Y}$ . Let the $\mathcal{P}_{in}$ be the marginal distribution of $\mathcal{X}$ which represents the distribution of input we want our classifier to be able to handle. This is the in-distribution (ID) of the input labels $x_i$.

\noindent\textbf{Supervised Classification}
In supervised classification, the goal is to minimize the empirical loss $\mathcal{L}$ function formulated as:
$\min_{\theta} \frac{1}{N} \sum_{i=1}^{N} \mathcal{L}(f_\theta(x_i), y_i)$ over the input dataset which is sampled \textit{i.i.d.} from the in-distribution $\mathcal{P}_{in}$. Here, $\theta$ is the model parameters, $f_\theta(x_i)$ is the classification predicted for input $x_i$ by the model with parameters $\theta$. 

\noindent\textbf{OOD Detection}
During test time the environment can present samples from a different distribution $\mathcal{P}_{out}$ instead of from $\mathcal{P}_{in}$. The goal of Out of Distribution Detection is to differentiate between samples from in-distribution $\mathcal{P}_{in}$ and out-of-distribution $\mathcal{P}_{out}$. In this work we treat OOD detection as a binary classification where a scoring function $S(\bx)$ and a corresponding threshold $\lambda$ provide a decision function the performs OOD detection: 
\begin{equation}
  g(\bx) = 
  \begin{cases}
    \text{In-distribution}, & \text{if } S(\bx) \geq \lambda \\
    \text{Out-of-distribution}, & \text{if } S(\bx) < \lambda
  \end{cases}
\end{equation}
The simplest of the scoring function $S(\bx)$ is the Maximum Softmax Probability (MSP) obtained by passing the logits from the final layer of the network to the softmax function and taking the maximum value. Then samples with MSP exceeding a certain threshold $\lambda$ are classified ID and the rest are OOD. The threshold $\lambda$ is usually chosen so as to have a true positive rate of 95\% over the input dataset.

\subsection{Motivation}
\begin{figure}[h]
  \centering
    \includegraphics[width=0.8\textwidth]{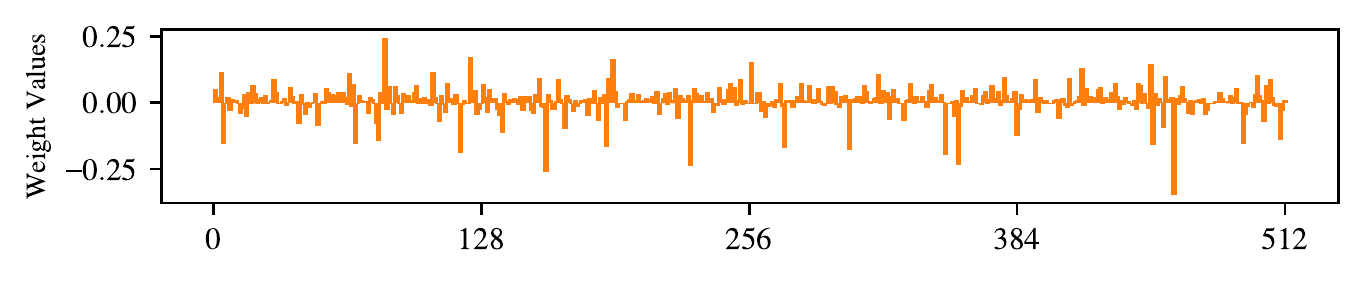}
    \caption{Smooth 
    FC weights of Airplane class in ResNet18 induced by LogitNorm\cite{wei2022mitigating} optimization.
    }
    \label{fig:fc_logitnorm}
\end{figure}

A recent work LogitNorm\cite{wei2022mitigating} directly aims to address the overconfidence issue by decoupling the effect of the norm of logits by L2 normalization in logits. Since the fully connected (FC) layer is directly responsible for logit computation and normalization is performed at logits, empirical observation (Figure \ref{fig:fc_logitnorm}) suggests that the optimization process induces smoother uniform weight values closer to zero in the FC layer. However, a recent work \cite{sun2021dice} also has shown that non-trivial dependence on unimportant weights (and units) can directly attribute to the brittleness of OOD detection. The presence of smoother weights 
implies irrelevant features contributing non-trivially to the classification for some predictions resulting in higher output variance for OOD samples. Furthermore, suppressing the logit norm by forcefully learning predominantly near-zero FC might only suboptimally address overconfidence at the feature level. Hence, we address the normalization in feature space to avoid unwanted implications on FC weights.

\subsection{Feature normalization} 

\begin{figure}[t]
  \centering
  \includegraphics[width=0.8\textwidth]{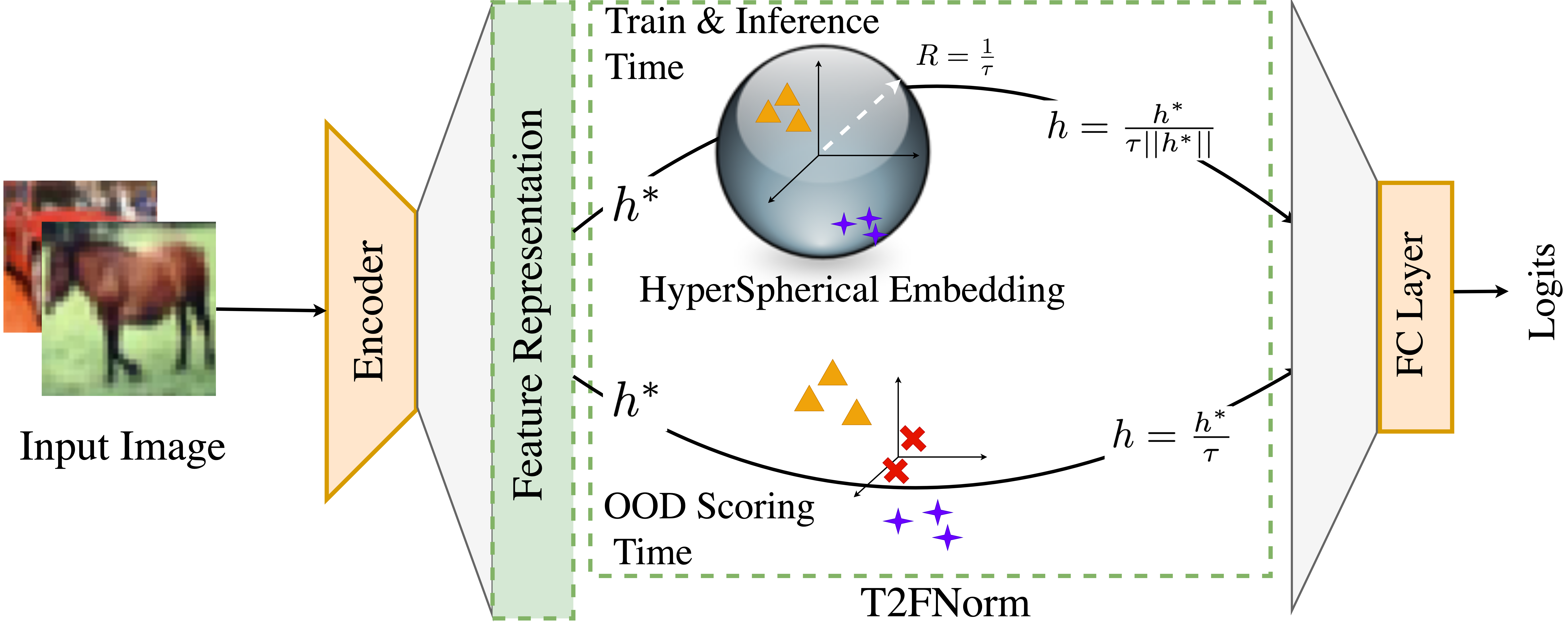}
   \caption{Schematic diagram of our method: T2FNorm. Features are L2 normalized and scaled during  training and inference time, while normalization is avoided for OOD Scoring.}\label{fig:schematic}
 \end{figure}

Our work proposes a method \textbf{T2FNorm} to improve the robustness of the network itself for OOD detection which can be used in conjunction with any downstream scoring function. We perform feature normalization to alleviate the issue of over-confidence predictions at the feature level. In particular, we normalize the feature vectors in the penultimate layer and scale with a factor $1/\tau$. The normalized vector is then passed on, as usual, to the classification FC layer and to cross-entropy loss function. Importantly, this normalization is performed (Algorithm \ref{alg:train}) only during training  and inference time, however, we skip the normalization part for OOD detection (Algorithm \ref{alg:infer}). Figure \ref{fig:schematic} shows the schematic diagram for our method. The proposed approach is simple and easy to implement, and as we will show later, it produces improved performance for OOD detection while maintaining predictive abilities. 
\begin{minipage}[t]{\textwidth}
\begin{minipage}[t]{0.45\textwidth}
\begin{algorithm}[H]
\begin{algorithmic}
   \STATE {\textbf{Input:}} Dataset $\mathcal{D}$, Feature Extractor $\phi$, \\ classification layer $FC$ 
   \FUNCTION{train($\mathcal{D}$)}
        \FOR{($\bx_i$, $\by_i$) $\gets$ $\mathcal{D}$}
            \STATE $h^* \gets$ $\phi(\bx_i)$
            \STATE $h \gets h^* / {\tau \lVert h^* \rVert_2}$
            \STATE $\mathcal{L} \gets$ cross\_entropy\_loss$(FC(h), \by_i)$
            \STATE $\mathcal{L}$.backward()
        \ENDFOR
    \ENDFUNCTION
\end{algorithmic}
\caption{T2Norm: Training}
\label{alg:train}
\end{algorithm}
\end{minipage}
\hfill
\begin{minipage}[t]{0.45\textwidth}
\begin{algorithm}[H]
\begin{algorithmic}
   \FUNCTION{classify($\bx$)}
        \STATE $h^* \gets \phi(\bx_i)$
      
        \IF{$S(\bx; h^*/\tau) < \gamma $} 
            \STATE return OOD
        \ELSE 
            \STATE $h \gets h^*/\tau \lVert h^* \rVert_2$
            \STATE logits $ \gets$ $FC(h)$
            \STATE return $argmax_i$ logits$_i$ 
        \ENDIF
    \ENDFUNCTION
\end{algorithmic}
\caption{T2FNorm: Inference}
\label{alg:infer}
\end{algorithm}
\end{minipage}
\end{minipage}

\noindent\textbf{Significance of Feature Norm} As observed by recent works \cite{tack2020csi, huang2021importance,igoe2022useful}, generally ID samples have a more significant penultimate feature norm in comparison to OOD data. In CNN models, high-level spatial features are generated by convolution operations. The penultimate feature is derived from globally pooling post-ReLU spatial features. ReLU activation signifies the presence of specific in-distribution features, while their absence corresponds to smaller norms, often seen in out-of-distribution samples. Therefore, a neural network having better ID/OOD separability should demonstrate a higher relative norm for in-distribution versus out-of-distribution samples, enhancing discriminability.

\noindent\textbf{Working Principle and Details} 
Our operational hypothesis is that the network learns to produce high-level semantic ID features lying on the hypersphere due to the normalization performed during training. However, this happens only for ID samples only as the network was trained with them, while for OOD samples, high-level 
semantic ID features are not activated because of their absence, causing OOD feature representation to lie significantly beneath the hypersphere's surface. The superior the degree to which this occurs, the greater the distinction between ID and OOD data samples occurs. Quantitatively, we can formalize such distinction as the ratio of ID norm to OOD norm, which we term as \textit{\textbf{separability ratio}} ($\mathcal{S})$ in this work.
 We observe that depending upon the \textit{separability ratio}, LogitNorm\cite{wei2022mitigating} and LogitPenalty\cite{wei2022mitigating} can perform OOD detection supporting the evidence of the preferability of a higher separability ratio.

As most out-of-distribution (OOD) detection metrics concentrate on logits, and prior research has primarily focused on these logits, we investigate the impact of normalization at the feature representation level on both the separability ratio and the norm in feature and logit spaces for OOD detection. Given that feature-level normalization implies normalization within a higher-dimensional space than logit-level normalization, we postulate that high-dimensional normalization of training ID samples would enable the network to significantly reduce OOD norm relative to ID norm while preserving ID-specific features. As a result, we anticipate a substantial decrease in overconfidence, which is intrinsically linked to logit and feature norms, primarily since overconfidence is addressed at the penultimate feature level, inherently tackling the norm at the logit level. Confirming this, recent work, ReAct\cite{sun2022out}, has observed that the penultimate layer is most effective for OOD detection due to the distinct activation patterns between ID and OOD data.

\textbf{On the significance of avoiding normalization at OOD scoring} Should feature normalization be adopted during OOD scoring, it erroneously activates the feature for OOD samples, causing them to mimic the behavior of ID samples within the feature space. But, the removal of normalization during OOD scoring helps to preserve the difference in response of the network towards OOD and ID samples in the feature space.

\section{Experiments}
In this section, we discuss the experiments performed in various settings to verify the effectiveness of our method.

\noindent\textbf{Datasets:}
We use CIFAR-10\cite{cifar-10} and CIFAR-100\cite{cifar-100} as in-distribution datasets. Texture\cite{texture}, TinyImageNet (TIN)\cite{imagenet12nips}, MNIST\cite{deng2012mnist}, SVHN\cite{svhn}, Places365\cite{zhou2017places}, iSUN\cite{xu2015turkergaze}, LSUN-r\cite{yu2015lsun}, LSUN-c\cite{yu2015lsun} are used as out-of-distribution datasets. Following \cite{yang2022openood}, we use CIFAR-10 as OOD if CIFAR-100 is used as in-distribution and vice versa.

\noindent\textbf{Metrics and OOD scoring:} We report the experimental results in three metrics: FPR@95, AUROC and AUPR. FPR@95 gives the false positive rate when the true positive rate is 95\%. AUROC denotes the area under the receiver operator characteristics curve and AUPR denotes the area under the precision-call curve. We use multiple OOD scoring methods, including parameter-free scoring functions such as maximum softmax probability\cite{hendrycks2016baseline}, parameter-free energy score \cite{liu2020energy} and GradNorm\cite{huang2021importance} as well as hyperparameter-based scoring functions such as ODIN\cite{liang2017enhancing} and DICE\cite{sun2021dice}. We use the recommended value of 0.9 for the DICE sparsity parameter $p$ and recommended $\tau=1000$ and $\epsilon=0.0014$ for ODIN.

\noindent\textbf{Training pipeline:}
We perform experiments with three training methods: a) Baseline (cross-entropy), LogitNorm \cite{wei2022mitigating}, and T2FNorm (ours) by following the training procedure of open-source framework OpenOOD\cite{yang2022openood}. Experiments were performed across ResNet-18, WideResnet(WRN-40-2), and DenseNet architectures with an initial learning rate of 0.1 with weight decay of 0.0005 for 100 epochs based on the cross-entropy loss function. We set the  temperature parameter $\tau=0.04$ for LogitNorm as recommended in the original setting\cite{wei2022mitigating} and $\tau=0.1$ for T2FNorm. 
Please refer to Figure \ref{fig:tau_ablation} for the sensitivity study of $\tau$. Five independent trials are conducted for each of 18 training settings (across 2 ID datasets, 3 network architectures, and 3 training methods). We trained all models on NVIDIA A100 GPUs.
\section{Results}\label{sec:results}

\begin{figure}[t]
  \begin{minipage}[t]{\linewidth}    
    \centering
    \includegraphics{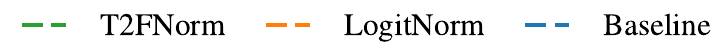}
  \end{minipage}
  \begin{minipage}[b]{0.45\linewidth} 
  \centering
  \includegraphics[width=0.8\linewidth]{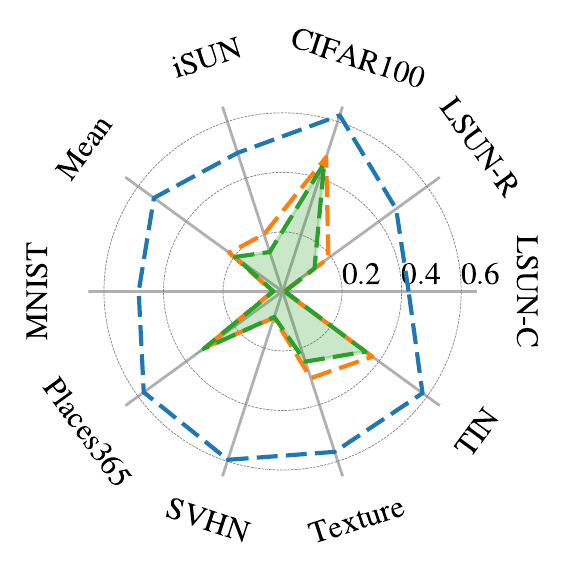}
      \caption{FPR@95 for CIFAR10 as ID (MSP)}\label{fig:radar_ood_cifar10_base_main}
  \end{minipage}
  \hfill
  \begin{minipage}[b]{0.45\linewidth} 
    \centering
      \includegraphics[width=0.8\linewidth]{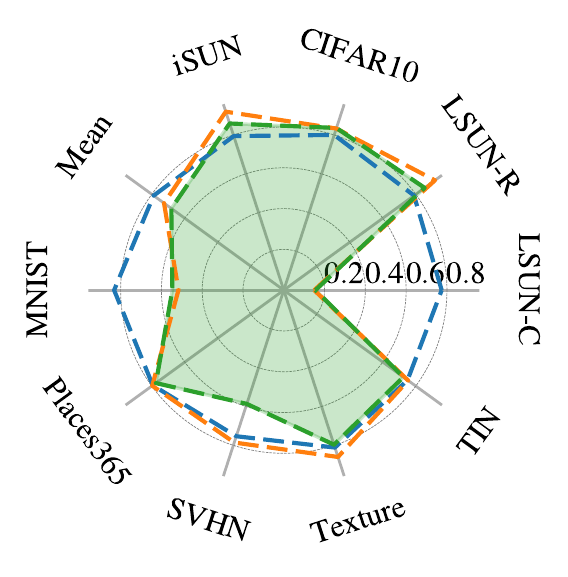}
    \caption{FPR@95 for CIFAR100 as ID (MSP)}\label{fig:radar_ood_cifar100_base_main}
    \end{minipage}
\end{figure}

\noindent\textbf{Superior OOD Detection Performance} Quantitative results are presented in Table \ref{tab:main_table}. It shows that our method is consistently superior in FPR@95, AUROC as well as AUPR metrics. Our method reduce FPR@95 metric by 34\% compared to Baseline and 7\% compared to LogitNorm  using DICE Scoring for ResNet-18. Figures \ref{fig:radar_ood_cifar10_base_main} and \ref{fig:radar_ood_cifar100_base_main} show the FPR@95 values across different OOD datasets using MSP scoring in ResNet-18 where our method reduces FPR@95 by 33.7\% compared to baseline and by 4.4\% compared to LogitNorm. Interestingly, for both ID datasets, we can also observe the incompatibility of LogitNorm with DICE scoring in DenseNet architecture where it underperforms even when compared to the baseline. On the other hand, our method is more robust regardless of architecture or OOD scoring method.

\begin{table}[h]
    \centering
    \caption{Mean OOD metrics in the form of Baseline/LogitNorm/T2FNorm with hyperparameter-free (MSP, EBO) as well as hyperparameter-based OOD scoring (DICE). \textbf{Bold} numbers are superior.}
    \label{tab:main_table}
\resizebox{0.98\linewidth}{!}{%
    \begin{tabular}{@{}l|l|ccc|ccc@{}}
        \toprule
         &          &  \multicolumn{3}{c|}{CIFAR-10} & \multicolumn{3}{c}{CIFAR-100} \\
      \cmidrule(lr){2-2} \cmidrule(lr){3-5} \cmidrule(lr){6-8}
         &  Network & FPR@95 $\downarrow$ & AUROC $\uparrow$ & AUPR$\uparrow$ & FPR@95$\downarrow$ & AUROC$\uparrow$ & AUPR$\uparrow$ \\
      \midrule
      \multirow{3}{*}{\begin{sideways}MSP\end{sideways}}
         & ResNet-18 & 53.4 / 22.1/ \textbf{19.7} & 90.7 / 96.0 / \textbf{96.5} & 90.8 / 95.7 / \textbf{96.4} & 78.9 / 72.6 / \textbf{68.2} & 79.0 / 80.1 / \textbf{83.2} & 79.8 / 79.4 / \textbf{82.4}\\ 
         & WRN-40-2 & 53.4 / 22.6 / \textbf{22.4} & 90.1 / 95.9 / \textbf{95.9} & 90.2 / 95.8 / \textbf{95.9} & 81.8 / 63.5 / \textbf{63.2} & 74.7/ 83.8 / \textbf{83.9} & 76.6 / 83.7 / \textbf{84.2}\\ 
         & DenseNet & 48.8 / 24.0 / \textbf{21.0} & 91.7 / 95.4 / \textbf{96.1} & 91.6 / 95.3 / \textbf{96.2} & 77.4 / 66.8 / \textbf{64.1} & 77.6 / 82.1 / \textbf{84.1} & 79.8 / 82.6 / \textbf{84.6}\\ \midrule
         & Mean & 51.9 / 22.9 / \textbf{21.0} & 90.9 / 95.8 /\textbf{96.2} & 90.9 / 95.6 / \textbf{96.2} & 79.4 / 67.6 /\textbf{65.1} & 77.1 / 82.0 /\textbf{83.7} & 78.7 / 81.9 / \textbf{83.7}\\ 
         \midrule
      \multirow{3}{*}{\begin{sideways}DICE\end{sideways}}
         & ResNet-18 & 54.5 / 27.6 / \textbf{20.5} & 86.0 / 94.4 / \textbf{96.3} & 87.4 / 94.1 / \textbf{96.1} & 76.7 / 68.6 / \textbf{64.9} & 81.0 / 73.5/ \textbf{83.1} & 81.2 / 74.6 / \textbf{81.9}\\ 
         & WRN-40-2 & 36.5 / 32.5 / \textbf{26.0} & 89.0 / 92.6 / \textbf{95.1} & 90.9 / 92.8 / \textbf{95.1} & 76.4/ 59.1 / \textbf{55.7} & 74.8 / 81.6 / \textbf{84.6} & 76.0 / 81.6 / \textbf{84.6}\\ 
         & DenseNet & 30.8 / 38.0 / \textbf{23.0} & 92.3 / 90.3 / \textbf{95.4} & 93.3 / 90.7 / \textbf{95.4} & 63.4/ 68.2 / \textbf{61.2} & 82.8 / 75.7 / \textbf{82.9} & \textbf{83.5} / 75.8 / 82.6\\
        \midrule
        & Mean & 40.6 / 32.7 / \textbf{23.2} & 89.1 / 92.5 / \textbf{95.6} & 90.5 / 92.6 / \textbf{95.5} & 72.2 / 65.3 / \textbf{60.6} & 79.6 / 76.9 / \textbf{83.5} & 80.2 / 77.3 / \textbf{83.0} \\ 
        \midrule
        \multirow{3}{*}{\begin{sideways}EBO\end{sideways}}
         & ResNet-18 & 37.7 / 37.0 / \textbf{17.9} & 91.5 / 88.9 / \textbf{96.7} & 92.7 / 89.4 / \textbf{96.6} & 77.6 / 72.6 / \textbf{66.6} & 81.0 / 75.1 / \textbf{83.3} & 81.2 / 75.3 / \textbf{82.2} \\ 
         & WRN-40-2 & 35.3 / 54.9 / \textbf{22.5} & 91.1 / 85.0 / \textbf{95.8} & 92.1 / 84.1 / \textbf{95.7} & 78.0 / 62.6 / \textbf{60.0}  & 77.0 / 81.7 / \textbf{84.2} & 78.3 / 81.9 / \textbf{84.4} \\ 
         & DenseNet & 30.3 / 73.9 / \textbf{20.0} & 93.3 / 86.3 / \textbf{96.1} & 93.8 / 83.2 / \textbf{96.1} & 69.2 / 70.3 / \textbf{62.2}  & 82.4 / 75.7 / \textbf{83.4} & 83.6 / 77.0 / \textbf{84.0} \\
        \midrule
        & Mean & 34.5 / 55.3 / \textbf{20.1} & 92.0 / 86.7 / \textbf{96.2} & 92.9 / 85.6 / \textbf{96.2} & 75.0 / 68.5 / \textbf{63.0} & 80.1 / 77.5 / \textbf{83.6} & 81.0 / 78.1 / \textbf{83.5} \\
        \bottomrule
    \end{tabular}
}
\end{table}


\noindent\textbf{Architecture Agnostic without Compromising Accuracy}
Our experiments across three architectures as reported in Table \ref{tab:main_table} show the compatibility of our method with various architectures evidencing the agnostic nature of our method to architectural designs. An essential attribute of OOD methods employing regularization during training is the preservation of classification accuracy in ID datasets, independent of their OOD detection performance. The evidence supporting these assertions can be found in Table \ref{tab:accuracy}.

\begin{table}
\small
\centering
\caption{Accuracy in \% with (Baseline / LogitNorm / T2FNorm)}\label{tab:accuracy}
\resizebox{0.5\linewidth}{!}{%
\begin{tabular}{ccc}
\toprule
Architectures & CIFAR-10 & CIFAR-100 \\ \midrule
DenseNet   & 94.94 / 94.05 / 94.62 & 76.51 / 76.50 / 76.06 \\
WRN-40-2 & 94.72 / 94.38 / 94.44 & 75.30 / 74.79 /  75.51 \\
ResNet-18  & 94.79 / 94.13 / 94.94 & 77.02 / 75.85 /  76.42 \\ \bottomrule
\end{tabular}
}
\end{table} 

\begin{figure}
  \centering
  \begin{subfigure}{0.34\textwidth}
    \includegraphics{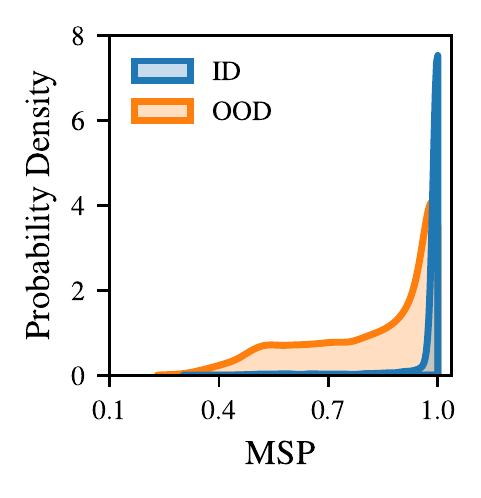}
    \caption{Baseline}
    \label{fig:figure_density_base}
  \end{subfigure}
  \hfill
  \begin{subfigure}{0.32\textwidth}
    \includegraphics{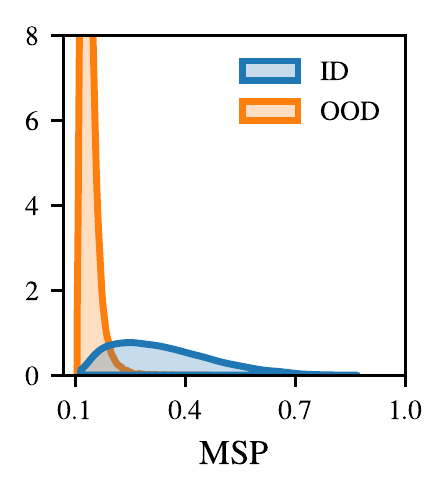}
    \caption{LogitNorm}
    \label{fig:figure4}
  \end{subfigure}
  \hfill
  \begin{subfigure}{0.32\textwidth}
    \includegraphics{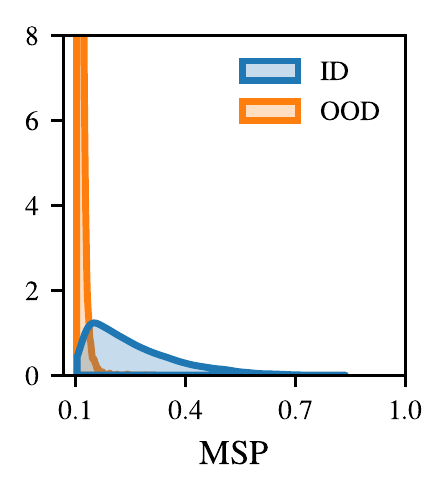}
    \caption{T2FNorm}
    \label{fig:figure_density_trainnorm}
  \end{subfigure}
  \caption{Distribution of Maximum Softmax Probability (MSP) shows that overconfidence is controlled in both T2FNorm and LogitNorm while the overlapping region is even smaller in T2FNorm.}
  \label{fig:msp_plot}
\end{figure}

\begin{figure}[h]
  \centering
  \begin{subfigure}{0.49\textwidth}
    \includegraphics{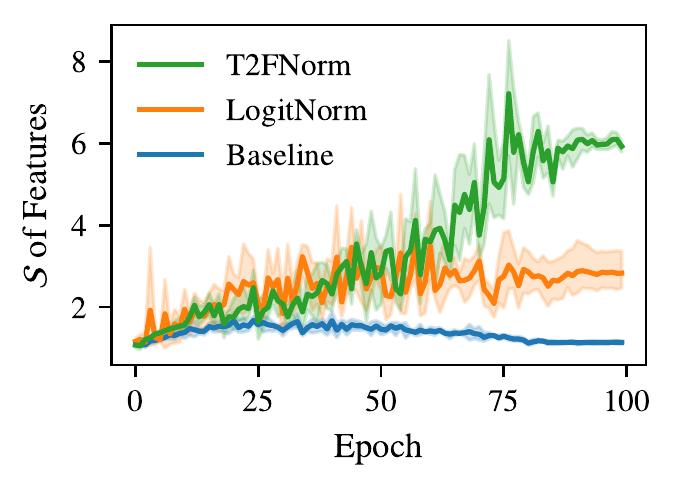}
  \end{subfigure}
  \hfill
  \begin{subfigure}{0.49\textwidth}
    \includegraphics{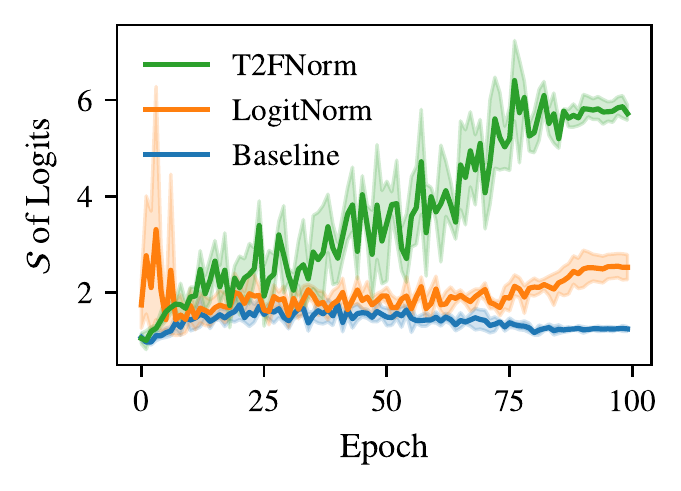}
  \end{subfigure}
  \caption{Progression of Separability Ratio $\mathcal{S}$ in feature and logit space over training epochs.}
  \label{fig:s_progression_cifar10} 
\end{figure}

\noindent\textbf{Significant Reduction in Overconfidence} In Figure \ref{fig:msp_plot} we show the comparison between Baseline, LogitNorm, and T2FNorm in terms of distribution of maximum softmax probability. It can be observed that overconfidence has been addressed by T2FNorm to a greater extent in comparison with the baseline. 
Though the issue of overconfidence is also reduced in LogitNorm, the separability ratio is significantly higher in T2Norm, as we show in Figures \ref{fig:s_progression_cifar10} and \ref{fig:s_progression_cifar100}.

\begin{table}[h]
  \small
  \centering
  \caption{Norm of features and logits for ID and OOD samples. (ID / OOD$\downarrow$ / $\mathcal{S}$ $\uparrow$)} 
  \resizebox{1\linewidth}{!}{%
    \begin{tabular}{ccccc}
      \toprule
      Method & Penultimate feature & Logit  \\
      \cmidrule(r){1-3}
      Baseline & 6.16 $\pm$ 0.27 / 5.43 $\pm$ 0.20 / 1.13 $\pm$ 0.05            & 10.25 $\pm$ 0.17 / 8.22 $\pm$ 0.38 / 1.25 $\pm$ 0.06            \\ 
      LogitNorm & 1.90 $\pm$ 0.11 / 0.69 $\pm$ 0.12 / 2.83 $\pm$ 0.58            &  1.98 $\pm$ 0.13 / 0.80 $\pm$ 0.14 / 2.53 $\pm$ 0.33            \\
      T2FNorm   & 0.90 $\pm$ 0.01 / 0.15 $\pm$ 0.00 / \textbf{6.01 $\pm$ 0.18}   &  1.02 $\pm$ 0.01 / 0.18 $\pm$ 0.01 / \textbf{5.78 $\pm$ 0.23}   \\
      \bottomrule
    \end{tabular}
  }\label{tab:separability_table} 
\end{table}


\paragraph{Norm and Separability Ratio}
The statistics of norm and separability ratio for ResNet-18 model trained with CIFAR-10 datasets are given in Table \ref{tab:separability_table}. The average ID norm of $0.9\sim1$ for the penultimate feature implies empirically that, ID samples approximately lie on the hypersphere even at the pre-normalization stage. Again, the average norm for OOD samples is found to be 0.15 implying OOD samples lie significantly beneath the hypersphere as ID-specific features are not activated appreciably. This depicts a clear difference in the response of the network towards OOD and ID. Similar observations can be found on logits as the feature representation has a direct implication on it. More importantly, from the comparison of various methods, we observe that the separability factor $\mathcal{S}$ induced by our method is highly significant. For instance, we achieve ($\mathcal{S} = 6.01$) at the end of training in the penultimate feature. The progression of S over the epochs in both the feature and logit space can be observed from Figure \ref{fig:s_progression_cifar10}. 

\begin{figure}
\centering
   \begin{minipage}[t]{0.49\linewidth} 
    \includegraphics[height=0.85\linewidth]{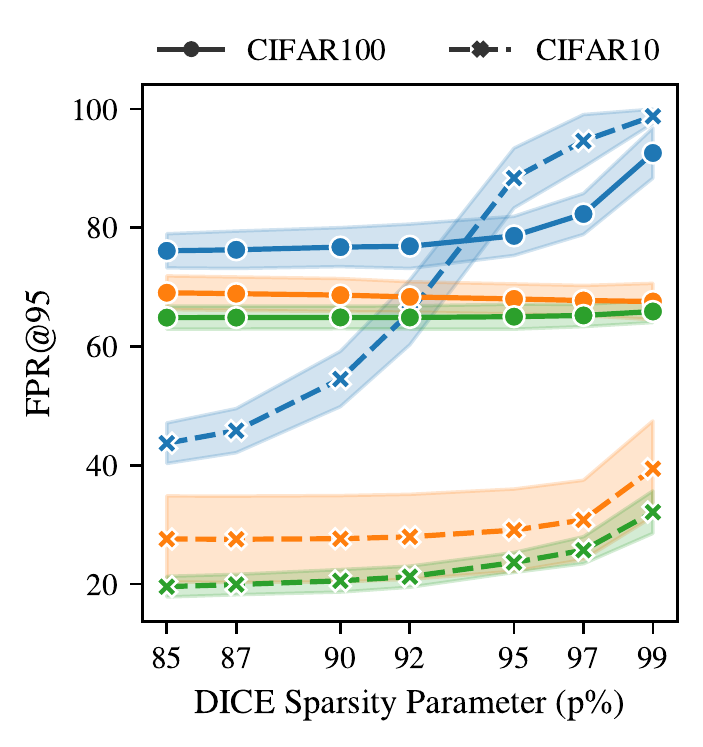}
     \centering
     \caption{Sensitivity study of sparsity parameter $p$ \cite{sun2021dice} shows superior performance and robustness of T2FNorm throughout the parameter range.}\label{fig:dice_sparsity}
    \end{minipage}
    \hfill
    \begin{minipage}[t]{0.49\linewidth}
     \includegraphics[height=0.85\linewidth]{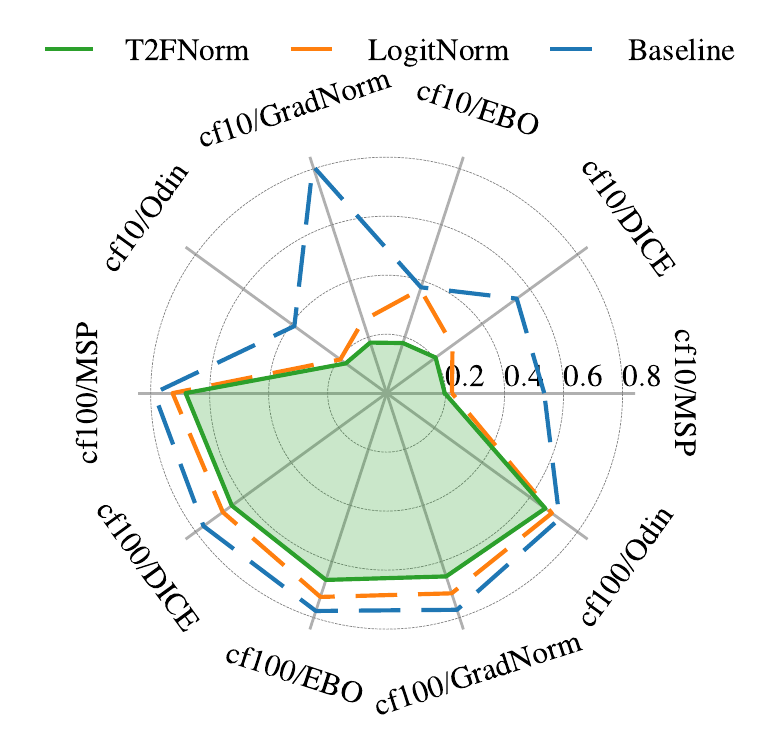}
    \centering
    \caption{FPR@95 across different scoring functions in CIFAR100 (cf100) and CIFAR10 (cf10) show superior performance of T2FNorm}\label{fig:radar_across_scoring}
    \end{minipage}
\end{figure}

\paragraph{Compatibility with existing OOD scoring methods}
T2FNorm is compatible with various existing OOD scoring functions. 
Figure \ref{fig:radar_across_scoring} shows that existing scoring functions when applied to the model trained with T2FNorm can boost the OOD detection performance. For instance, our model improves the baseline's OOD performance using ODIN from FPR@95 of 38.67 to  17.15 in ResNet18 architecture for CIFAR-10 experiments. Hyperparameter-free energy-based scoring function can also get a boost of 19.86 
in comparison to the baseline model. Similarly, DICE \cite{sun2021dice} works very well (Figure \ref{fig:dice_sparsity}) with our method too.

\section{Discussion}
\begin{figure}[h]
  \centering
  \begin{minipage}[t]{0.45\textwidth}
    \includegraphics[height=140pt]{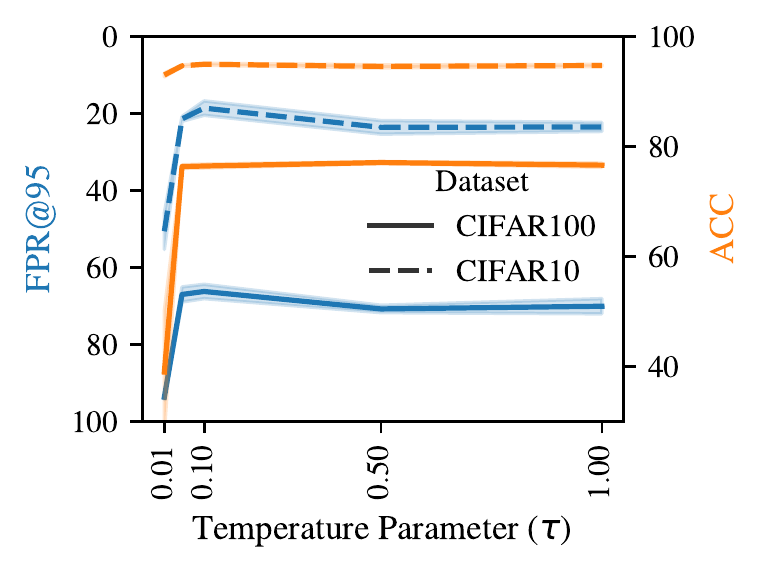}
    \caption{Sensitivity study of temperature $\tau$
    }
    \label{fig:tau_ablation}
  \end{minipage}
  \hfill
  \begin{minipage}[t]{0.45\textwidth}
    \includegraphics[height=140pt]{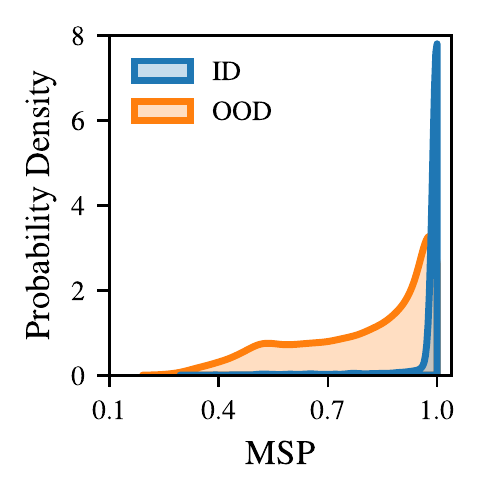}
    \caption{Normalization at OOD scoring}
    \label{fig:test_normalization}
  \end{minipage}
\end{figure}

\paragraph{Ablation Study of Normalization}\label{sec:ablation_of_normalization}
Imposing normalization during OOD scoring enforces a constant magnitude constraint on all inputs, irrespective of their originating distribution. This effectively eradicates the very characteristic (the magnitude property) that could potentially differentiate whether an input originates from the training distribution or not. It results in the trained network incorrectly assuming OOD samples as ID samples. As demonstrated in Figure \ref{fig:test_normalization}, the separability of the nature of input distribution is compromised by normalization during OOD scoring. Quantitatively, for trained ResNet-18 architecture with CIFAR-10 as ID, this degrades the mean FPR@95 performance from 19.7\% (T2FNorm) to 48.66\%.

\paragraph{Sensivity Study of Temperature $\tau$} 
Figure \ref{fig:tau_ablation} shows that the classification accuracy and OOD Detection performance (FPR@95) are not much sensitive over a reasonable range of $\tau$. We found the optimal value of $\tau$ to be 0.1. And while the performance is good for $\tau \in [0.05, 1]$, both accuracy and FPR@95 score degrades substantially for $\tau > 1$ and $\tau \leq 0.01$.

\begin{figure}[h]
  \centering
    \includegraphics[width=0.7\textwidth]{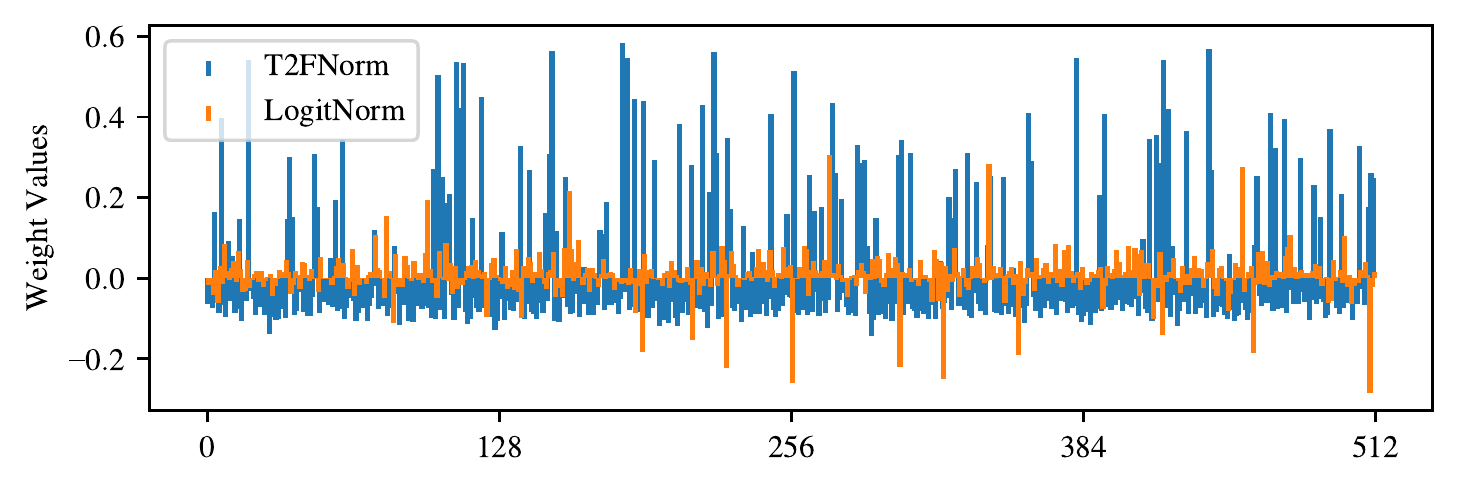}
    \caption{FC layer's weight comparison of Airplane class.}
    \label{fig:fc_comparison}
\end{figure}

\paragraph{Implication on FC Layer Weights} Figure \ref{fig:fc_comparison} shows the weights of the final classification layer corresponding to the Airplane class for T2FNorm and LogitNorm. In comparison to the smoother weight of LogitNorm, weights of T2Norm have higher variance and are sharply defined. Quantitatively, we find the average variance to be about 10 times higher in T2FNorm as compared to LogitNorm. Roughly speaking, it can be inferred that T2FNorm encourages the clear assignment of important features for a given category classification. It necessitates the activation of the specific important features for ID sample predictions. Conversely, OOD samples, which lack these important features, fail to activate them, leading to lower softmax probabilities.  
Table \ref{tab:fc_weight_statistics} also further shows that the mean of both the negative weights and positive weights are greater in magnitude for T2FNorm.

\begin{table}[h]
    \centering
    \caption{Mean of the FC Layer weights for a single class and for all classes shows that T2FNorm has more distinctly assigned weights}
    \label{tab:fc_weight_statistics}
    \resizebox{1\linewidth}{!}{%
    \begin{tabular}{@{}l|ccc|ccc@{}}
        \toprule
        & \multicolumn{3}{c|}{Mean weights of Airplane Class} & \multicolumn{3}{c}{ Mean weights of All Class} \\
        \cmidrule(lr){1-2} \cmidrule(lr){2-4} \cmidrule(lr){5-7}
        Method & All Weights & Negative Weights & Positive Weights & All Weights & Negative Weights & Positive Weights \\
        \midrule
        Baseline & 0.000 & -0.062 & 0.102 & 0.000 & -0.056 & 0.107 \\
        LogitNorm & 0.007 & -0.042 & 0.027 & -0.003 & -0.027 & 0.032 \\
        T2FNorm & 0.005 & \textbf{-0.075} & \textbf{0.261} & 0.000 & \textbf{-0.072} & \textbf{0.283} \\
    \bottomrule
    \end{tabular}
    }
\end{table}

\section{Related Works}

\paragraph{OOD Detection}
Numerous studies have emerged in recent years focusing on OOD detection. A straightforward method for OOD detection is a simple maximum softmax probability \cite{hendrycks2016baseline}. However, it remains an unreliable scoring metric for OOD detection because of inherent overconfidence imposed by training with one-hot labels\cite{nguyen2015deep}. OOD detection has been primarily tackled with three lines of approach in the literature
(a) post-hoc methods, (b) outlier exposure and (c) train-time regularization.Post-hoc methods \cite{hendrycks2016baseline,haoqi2022vim,liu2020energy,sun2021dice,sun2021react,species22icml,mahananobis18nips,gram20icml,djurisic2023extremely} aim to improve the ID/OOD separability with pretrained models trained only with the aim for accuracy. 
Outlier exposure is another less studied line in academic research, as the assumption of the nature of OOD limits the ideal applications. However, it is found to be commonly used for industrial purposes.
Training time regularization\cite{lee2017training, bevandic2018discriminative, wei2022open, ming2022poem, hendrycks2018deep, katz2022training} employs some form of regularizer in the training scheme, and this line of work due to its capacity to directly impose favorable 
constraints during training potentially offers the most promising path to superior performance for OOD detection. For instance, LogitNorm \cite{wei2022mitigating} employs logit normalization as training time regularization to address the overconfidence issue and, thereby, improve OOD detection. Furthermore, LogitNorm \cite{wei2022mitigating} shows overconfidence can somewhat be addressed sub-optimally with logit penalty too. Different from LogitNorm \cite{wei2022mitigating}, our work pertains to addressing overconfidence in the feature space thereby automatically addressing overconfidence in the logit space. Needless to say, our work deals with high-dimensional normalization.

\paragraph{Normalization}
The utility of normalization in ensuring consistent input distribution and reducing covariate shift has proven beneficial in various subareas of deep learning\cite{sohn2016improved, wu2018unsupervised, zhang2019adacos, ranjan2017l2}. Normalization consisting of learnable parameters such as Batch Normalization\cite{ioffe2015batch}, Layer Normalization\cite{ba2016layer}, and Group Normalization\cite{wu2018group}, have been effective in mitigating training issues of neural networks. On the other hand, the strategic placement of L2 normalization has also been a popular recipe for training more effective deep learning models. Similar to our work, \cite{ranjan2017l2} constrains the features to lie on the hypersphere of fixed radius for face verification purposes but does so in both the training and testing phase without scaling. Further works in deep metric learning such as  ArcFace \cite{deng2019arcface}, CosFace \cite{wang2018cosface}, SphereFace \cite{liu2017sphereface}, etc realize the effectiveness of normalization. Specifically, \cite{cosinesim20accv} shows the hyperparameter-free OOD detection method introducing cosine loss by taking inspiration from norm face \cite{wang2017normface} where both the penultimate feature and fully connected layer are normalized. Our approach differs from cosine loss in three different ways. a) The temperature parameter is learned in the cosine loss method whereas we set a fixed temperature across all 6 settings. While it may seem extra hyperparameter is being added, we find a value of $\tau$ being architecture agnostic as well as dataset agnostic. b) Unlike cosine loss, we avoid normalizing the classification layer freeing it to learn non-smooth weight values which, in turn, boost compatibility with various downstream OOD scoring methods as they rely on ID-OOD separability based on magnitudes. c) Importantly, we remove the constraint of hyperspherical embeddings in the OOD scoring phase while \cite{cosinesim20accv} uses cosine similarity and is not compatible with other OOD scoring functions. \cite{guo2017calibration} provided a study showing modern neural networks' poor calibration and proposed to use temperature scaling as posthoc method to improve calibration. Platt scaling \cite{platt1999probabilistic} is another simple postprocessing calibrating technique. Label smoothing \cite{szegedy2016rethinking} helps to avoid overconfident calibration by adding uncertainty to the one-hot encoding of labels. 
\section{Conclusion}
In summary, our work introduces a novel training-time regularization technique, termed as \textbf{T2FNorm}, which seeks to mitigate the challenge of overconfidence via enhancing ID/OOD separability. We empirically show that T2FNorm achieves a higher separability ratio than prior works. This study delves into the utility of feature normalization to accomplish this objective. Notably, we apply feature normalization exclusively during the training and inference phases, deliberately omitting its application during the OOD scoring process. This strategy improves OOD performance across a broad range of downstream OOD scoring metrics without impacting the model's overall accuracy. We provide empirical evidence demonstrating the versatility of our method, establishing its effectiveness across multiple architectures and datasets. We also empirically show our method is less sensitive to the hyperparameters.
\section{Broader Impact and Limitations}
OOD detection is a crucial task regarding AI safety. The accuracy of OOD detection directly impacts the reliability of many AI applications. Safe deployment of AI applications is crucial in areas such as healthcare and medical diagnostics, autonomous driving, malicious use or intruder detection, fraud detection, and others. In such cases, OOD detection can play a crucial role in identifying and increasing robustness against unknown samples. Further, OOD detection also helps in increasing the trustworthiness of AI models to increase public acceptance of them. We demonstrate versatility across multiple datasets and architecture; however, due to the limited availability of compute, our experiments are limited to smaller resolution images from CIFAR-10 and CIFAR-100, and the results as such can't be guaranteed to generalize for higher resolution images or in real-life scenarios.
\section{Acknowledgement}
This work was supported by the Wellcome/EPSRC Centre for Interventional and Surgical Sciences (WEISS) [203145Z/16/Z]; Engineering and Physical Sciences Research Council (EPSRC) [EP/P027938/1, EP/R004080/1, EP/P012841/1]; The Royal Academy of Engineering Chair in Emerging Technologies scheme; and the EndoMapper project by Horizon 2020 FET (GA 863146).

\bibliographystyle{unsrt}
\bibliography{main}

\clearpage
\appendix
{\huge Appendix}
\section{Optimality of L2 normalization}
In addition to L2 normalization, we investigated various other normalization types, including L1, L3, and L4. While these alternative forms of normalization also enhance performance, L2 emerges as the most effective. As demonstrated in Table \ref{tab:normalization_type}, all the investigated normalizations outperform the baseline, underscoring the efficacy of normalization in general. Though the separability factor in feature space for L1 normalization is higher, the FPR@95 for L2 normalization is still superior along with a higher separability factor in logit space.

\begin{table}[ht]
\centering
\caption{L2 normalization proves to be the optimal form of normalization for T2FNorm}
\begin{tabular}{lccccc}
\toprule
  $L_{p}$ &  FPR@95\% &  AUROC &  AUPR &  $\mathcal{S}$ (Logit) &  $\mathcal{S}$ (Feature) \\
\midrule
  $p=1$ &    24.0 &   95.8 &  95.9 &      3.5 &        \textbf{8.5} \\
  $p=2$ &    \textbf{19.7} &   \textbf{96.5} &  \textbf{96.4} &      \textbf{6.01} &       5.8 \\
  $p=3$ &    20.8 &   96.4 &  96.3 &      5.0 &        4.7 \\
  $p=4$ &    20.1 &   96.4 &  96.3 &      5.3 &        4.6 \\
\bottomrule
\end{tabular}
\label{tab:normalization_type}
\end{table}

\section{Adverse effect of OOD scoring-time normalization}
As previously discussed in Section \ref{sec:ablation_of_normalization}, the utilization of OOD scoring-time normalization gives rise to a detrimental consequence primarily due to the obfuscation of the inherent distinction between ID and OOD samples in terms of their magnitudes. Using the ResNet-18 model trained with CIFAR-10, the results on various OOD datasets are more clearly summarized in Table \ref{tab:adverse_effect_test_normalization_cifar10} with three OOD metrics.
\begin{table}[ht] 
  \centering
  \caption{OOD metrics with ResNet-18 model trained in CIFAR-10 datasets for the comparison of OOD scoring-time Normalization Adaptation and OOD scoring-time Normalization Avoidance in the form of Adaptation/Avoidance. We consistently observe the superior  gain in OOD detection performance thereby validating the avoidance of normalization at scoring time.
  }
  \begin{tabular}{cccc}
    \toprule
    Datasets & FPR@95\% &  AUROC & AUPR \\
    \midrule
    CIFAR-100 & 60.8 / \textbf{45.3} & 88.9 / \textbf{91.6} & 86.0 / \textbf{90.2} \\
    TIN & 56.6 / \textbf{34.4} & 90.6 / \textbf{94.2} & 87.6 / \textbf{93.0} \\
    MNIST & 44.1 / \textbf{03.1} & 94.1 / \textbf{99.3} & 98.8 / \textbf{99.9} \\
    SVHN & 50.4 / \textbf{09.0} & 93.7 / \textbf{98.3} & 96.4 / \textbf{99.3} \\
    Texture & 48.4 / \textbf{24.8} & 92.9 / \textbf{95.6} & 85.7 / \textbf{93.0} \\
    Places365 & 56.5 / \textbf{32.8} & 90.8 / \textbf{94.3} & 96.5 / \textbf{98.2} \\
    iSUN & 48.2 / \textbf{13.9} & 93.2 / \textbf{97.6} & 90.6 / \textbf{97.1} \\
    LSUN-c & 27.3 / \textbf{00.9} & 96.2 / \textbf{99.8} & 95.7 / \textbf{99.8} \\
    LSUN-r & 45.6 / \textbf{13.3} & 93.8 / \textbf{97.7} & 92.1 / \textbf{97.5} \\
    \bottomrule
  \end{tabular}\label{tab:adverse_effect_test_normalization_cifar10}
\end{table}

\section{Feature norm Penalty}
Suppressing the norm, which is directly linked to overconfidence \cite{wei2022mitigating}, can also be feasible by employing the L2 norm of feature as the additional regularization loss. Using the joint optimization of $ L_{S} + \lambda L_{FP} $ ($L_{S}$ referring to supervised loss, $L_{FP}$ referring feature norm penalty loss), setting $\lambda=0.01$ to not affect the accuracy, we indeed find little improvement over baseline in FPR@95 metric only. The optimization objective seems to be satisfied with the relatively smaller average norm for both ID and OOD. However, the desired ID/OOD separability is not quite achieved as seen in \ref{tab:feature_penalty_comparison}.

\begin{table}[ht]
\centering
\caption{Comparison of Baseline and Feature norm penalty method}
\begin{tabular}{cccccc}
\toprule
Method & FPR@95\% &  AUROC & AUPR & ID norm & OOD norm \\
\midrule
Baseline & 53.4 & 90.7 & 90.8 & 11.8 & 10.3 \\
Feature norm penalty & 50.9 & 89.1 & 90.6 & 3.0  & 2.8 \\
\bottomrule
\end{tabular}
\label{tab:feature_penalty_comparison}
\end{table}

\section{Ablation study on Different Layers}
The ResNet-18 architecture is primarily comprised of four residual blocks: Layer1, Layer2, Layer3, and Layer4. We conducted an ablation study to understand the impact of the application of T2FNorm on representation obtained after each of these layers and discovered that pooled representation obtained from Layer4 was the only effective way to boost OOD detection performance \ref{tab:normalization_ablation}. This finding aligns with observations documented in \cite{sun2021react} where it was noted high-level features are considered to have substantial potential for distinguishing between ID and OOD data as the earlier layers primarily handle low-level features, while the later layers process semantic-level features. Furthermore, as we note in \ref{tab:normalization_ablation}, implementing normalization in the earlier layers results in a performance that is even poorer than the baseline.

\begin{table}[ht]
\centering
\caption{Impact of Feature Normalization in the $n^{th}$ Layer}
\begin{tabular}{cccc}
\toprule
Feature Normalization in $n^{th}$ Layer &  FPR@95\% &  AUROC &  AUPR \\
\midrule
Layer 1           &    96.6 &   43.2 &  55.8 \\
Layer 2           &    95.1 &   51.7 &  59.4 \\
Layer 3           &    80.7 &   76.6 &  78.2 \\
Layer 4           &   \textbf{19.7} &   \textbf{96.5} &  \textbf{96.4} \\ 
\midrule
Baseline          &  53.43  &  90.74  &  90.83 \\
\bottomrule
\end{tabular}
\label{tab:normalization_ablation}
\end{table}

\section{FC Weights Visualization}

\begin{figure}
\centering
    \begin{subfigure}[t]{0.45\linewidth}
    \centering
     \includegraphics[height=0.9\linewidth]{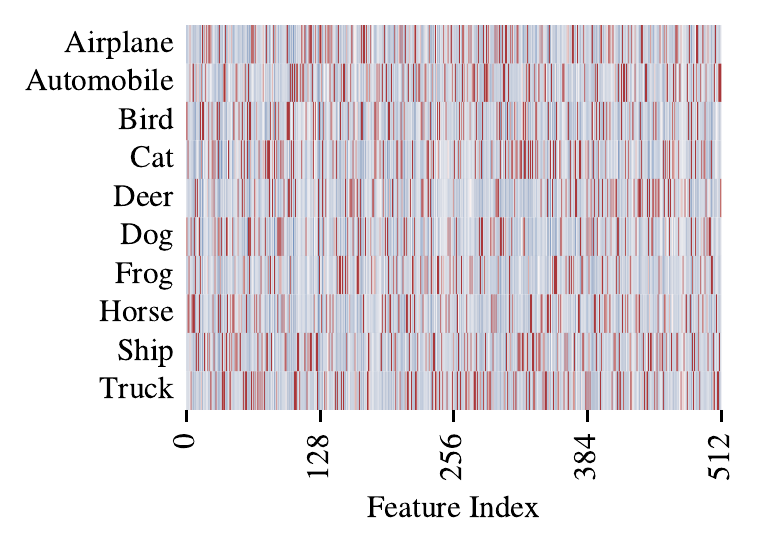}
     \caption{T2FNorm}\label{fc_visualization_T2FNorm_resnet}
    \end{subfigure}
    \hfill
    \begin{subfigure}[t]{0.45\linewidth}
       \centering
     \includegraphics[height=0.9\linewidth]{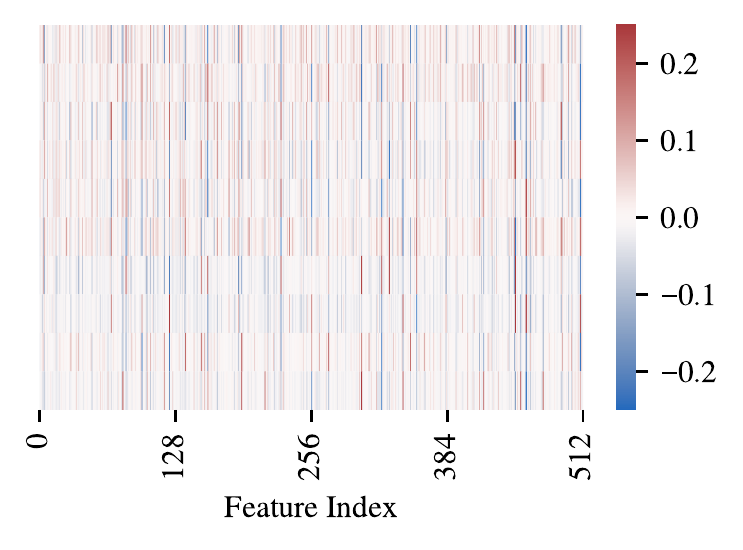}
     \caption{LogitNorm}\label{fc_visualization_LogitNorm_resnet}
    \end{subfigure}
    \caption{FC Weights heatmap in ResNet-18}
    \label{fig:fcweights}
\end{figure}

\begin{figure}
\centering
    \begin{subfigure}[t]{0.45\linewidth}
     \includegraphics[height=0.9\linewidth]{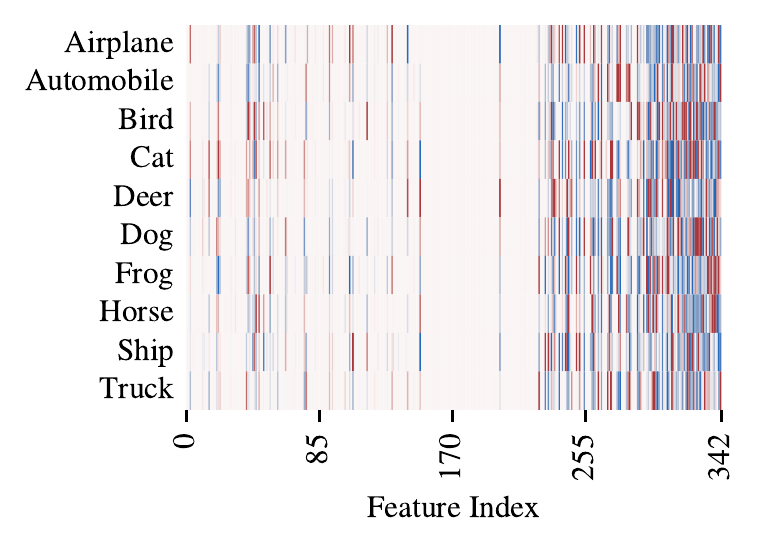}
     \centering
     \caption{T2FNorm}\label{fc_visualization_T2FNorm_densenet}
    \end{subfigure}
    \hfill
    \begin{subfigure}[t]{0.45\linewidth}
     \includegraphics[height=0.9\linewidth]{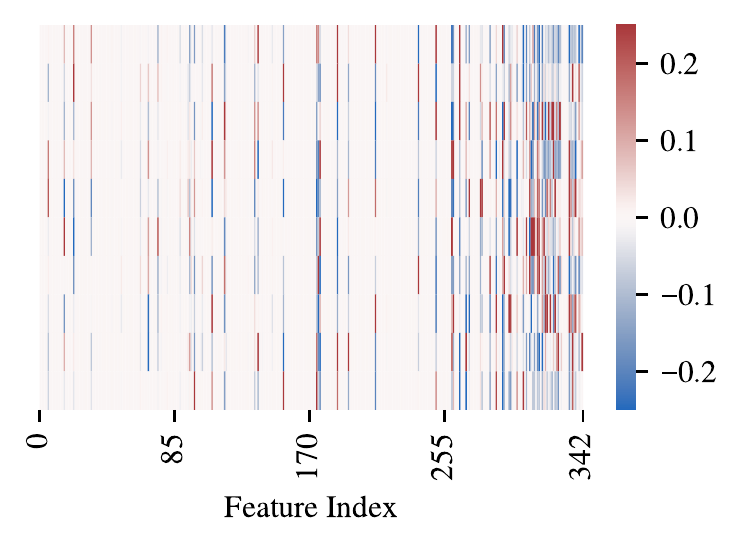}
     \centering
     \caption{LogitNorm}\label{fc_visualization_LogitNorm_densenet}
    \end{subfigure}
    \caption{FC Weights heatmap in DenseNet}
    \label{fig:fcweights_densenet}
\end{figure}

\begin{figure}
\centering
    \begin{subfigure}[t]{0.45\linewidth}
     \includegraphics[height=0.9\linewidth]{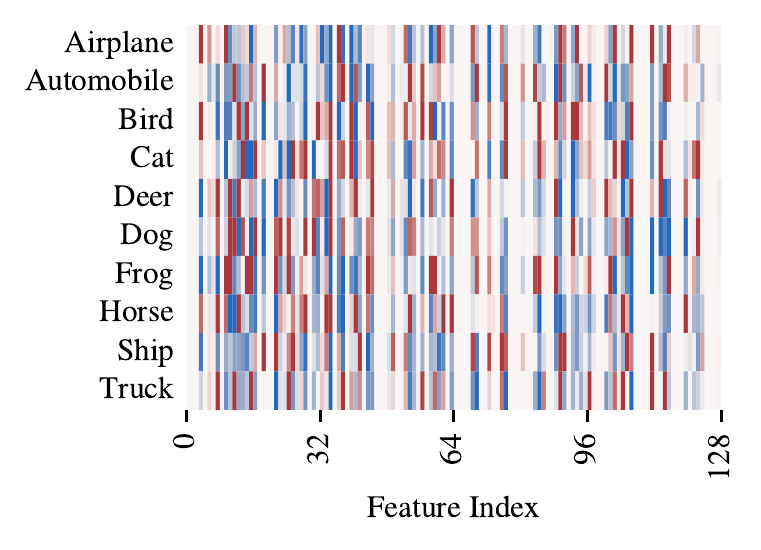}
     \centering
     \caption{T2FNorm}\label{fc_visualization_T2FNorm_wrn}
    \end{subfigure}
    \hfill
    \begin{subfigure}[t]{0.45\linewidth}
     \includegraphics[height=0.9\linewidth]{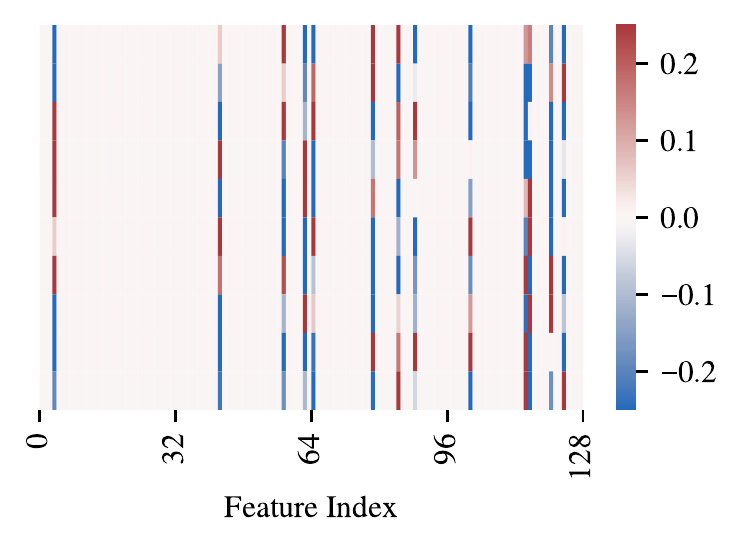}
     \centering
     \caption{LogitNorm}\label{fc_visualization_LogitNorm_wrn}
    \end{subfigure}
    \caption{FC Weights heatmap in WRN-40-2}
    \label{fig:fcweights_wrn}
\end{figure}

We show weight visualization of all the classes of fully connected layers in Figure \ref{fig:fcweights}, \ref{fig:fcweights_densenet} and \ref{fig:fcweights_wrn} for both T2FNorm as well as LogitNorm across various architectures. It is clearly evident that LogitNorm induces smaller weights on the FC layer in comparison to T2FNorm from the observation of all corresponding 10 classes of CIFAR-10 datasets. The observations show feature importance is sharper for our method in comparison to LogitNorm.

\section{Norm and Separability ratio statistics (for CIFAR-100 as ID)}
The statistics of norms in both feature space and logit space along with the separability ratio obtained from the ResNet-18 model trained in CIFAR-100 datasets with various methods are given in Table \ref{tab:separability_table_c}. The observation is similar to the earlier observation, CIFAR-10 as ID. For instance, the separability ratio achieved with our method in the penultimate feature is 2.9 with a significantly lower norm of 0.32 for OOD data in comparison to other methods. We use SVHN as OOD data for the purpose of illustrating the statistics in all settings unless otherwise noted.


\begin{table}[h!]
\centering
\caption{Norm of features and logits for ID and OOD samples. (ID / OOD$\downarrow$ / S ratio $\uparrow$)}
\resizebox{0.55\linewidth}{!}{%
\begin{tabular}{ccc}
\toprule
Method & \multicolumn{1}{c}{Penultimate feature} & \multicolumn{1}{c}{Logit} \\
\midrule
Baseline  & 11.83 /  10.32 / 1.15    & 18.47 / 14.71 / 1.26 \\
LogitNorm & 1.63 / 1.21  / 1.37     &  1.03 /   0.66 / 1.57 \\
T2FNorm   & 0.90 /   0.32 / \textbf{2.88} &   1.31 / 0.47 / \textbf{2.80} \\
\bottomrule
\end{tabular}
}
\label{tab:separability_table_c}
\end{table}

\begin{figure}
  \centering
  \begin{subfigure}{0.49\textwidth}
    \includegraphics{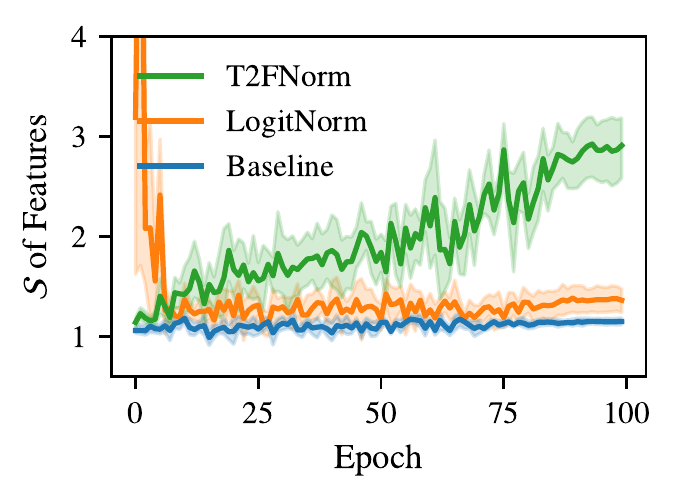}
  \end{subfigure}
  \hfill
  \begin{subfigure}{0.49\textwidth}
    \includegraphics{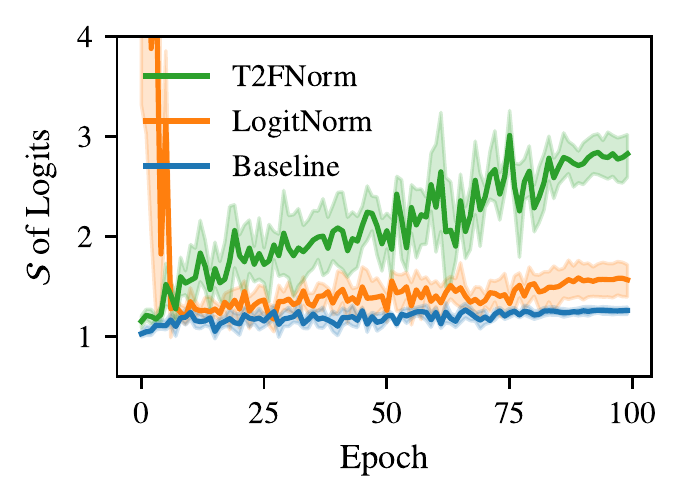}
  \end{subfigure}
  \caption{Progression of Separability Ratio $\mathcal{S}$ in feature and logit space over training epochs for CIFAR-100.}
  \label{fig:s_progression_cifar100}
\end{figure}

\section{Progression of ID and OOD norm with epochs (CIFAR 10)}
\begin{figure}
  \centering
  \begin{subfigure}{\textwidth}
      \centering
      \includegraphics{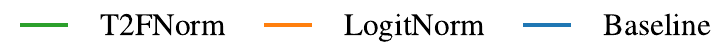}
  \end{subfigure}
  \begin{subfigure}{0.45\textwidth}
    \includegraphics{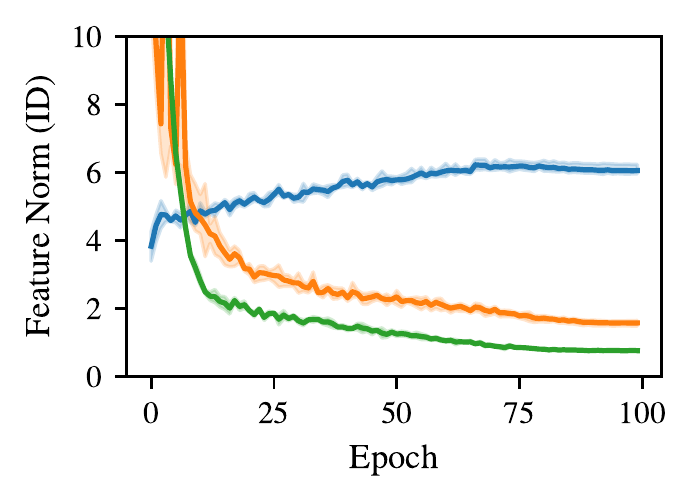}
  \end{subfigure}
  \hfill
  \begin{subfigure}{0.45\textwidth}
  \includegraphics{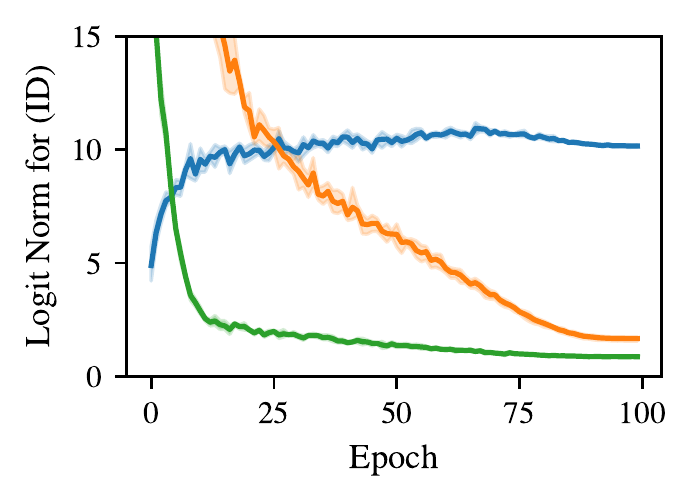}
  \end{subfigure}

   \begin{subfigure}{0.45\textwidth}
    \includegraphics{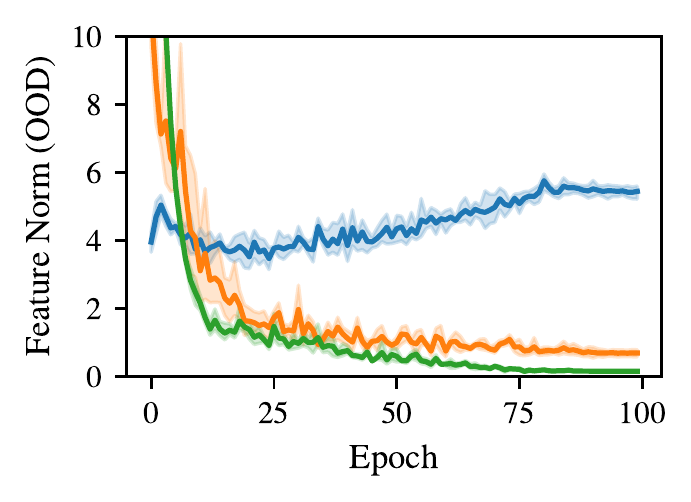}
  \end{subfigure}
  \hfill
  \begin{subfigure}{0.45\textwidth}
  \includegraphics{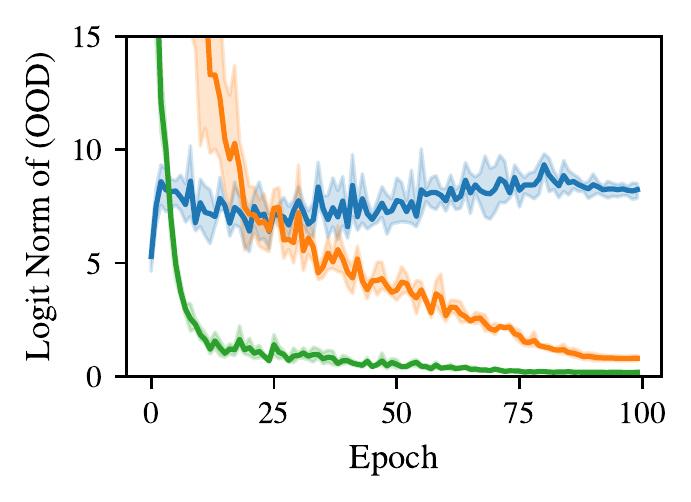}
  \end{subfigure}
  
  \caption{Progression of Norm in feature and logit space over training epochs (CIFAR10).}\label{fig:norm_progression_cifar10}
\end{figure}

\begin{figure}
  \centering
  \begin{subfigure}{\textwidth}
      \centering
      \includegraphics{figures/Norm_progress/legend.pdf}
  \end{subfigure}
  \begin{subfigure}{0.45\textwidth}
    \includegraphics{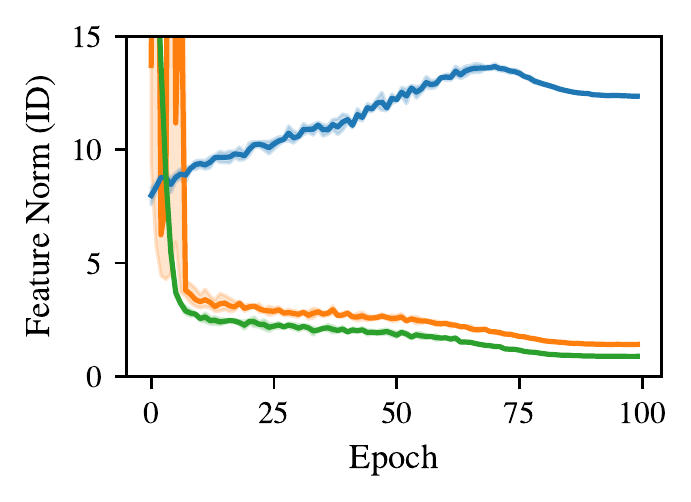}
  \end{subfigure}
  \hfill
  \begin{subfigure}{0.45\textwidth}
  \includegraphics{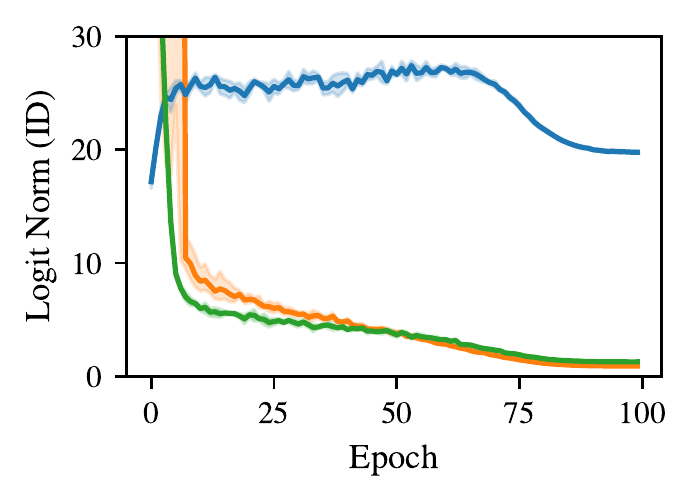}
  \end{subfigure}

  \begin{subfigure}{0.45\textwidth}
    \includegraphics{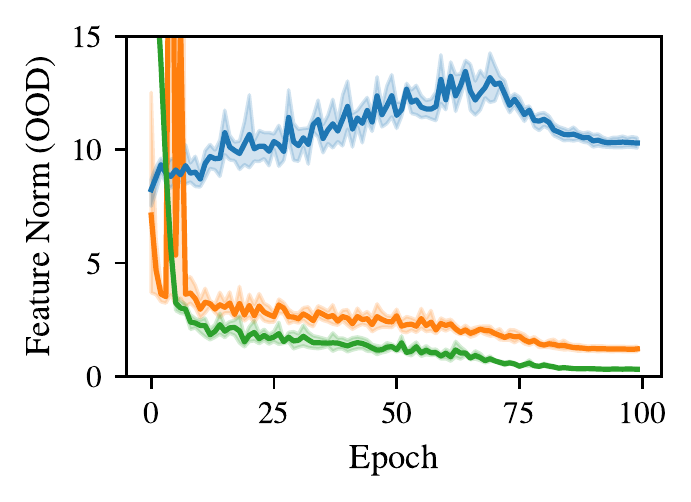}
  \end{subfigure}
  \hfill
  \begin{subfigure}{0.45\textwidth}
  \includegraphics{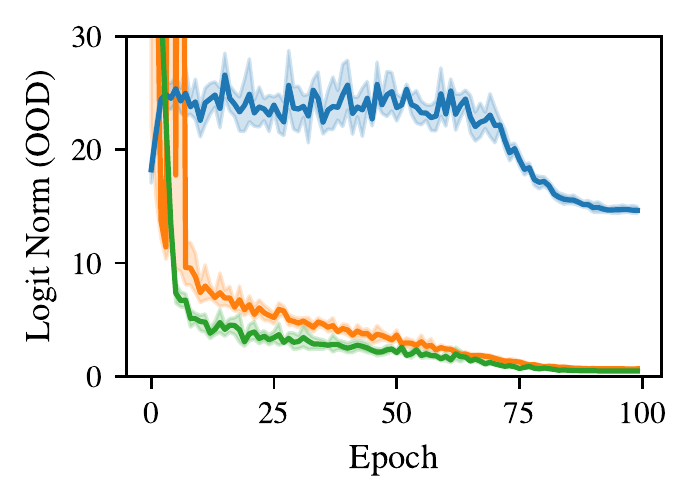}
  \end{subfigure}
  
  \caption{Progression of Norm $\mathcal{S}$ in feature and logit space over training epochs (CIFAR100).}\label{fig:norm_progression_cifar100}
\end{figure}

We show (Figure \ref{fig:norm_progression_cifar10}, \ref{fig:norm_progression_cifar100}) the progression of the average ID norm and average OOD norm in both feature and logit space with epochs during the training of ResNet-18 models. 

\subsection{Datasets}
\subsubsection{ID datasets}
CIFAR-10 and CIFAR-100 are two ID datasets used in our experiments.

\paragraph{CIFAR-10}
CIFAR-10 is one of the most commonly used datasets for benchmarking computer vision performance, especially for classification tasks. It contains 10 categories of images.
\paragraph{CIFAR-100}
CIFAR-100 is a very similar dataset to CIFAR-10 but consists of 100 classes.

\subsubsection{OOD datasets}
We use a total of 9 OOD datasets: MNIST, iSUN, CIFAR (CIFAR-10 for CIFAR-100 as ID and CIFAR-100 for CIFAR-10 as ID), TinyImagenet (TIN), LSUN-R, LSUN-C, Places365, SVHN, and Texture.
\paragraph{MNIST}
The MNIST dataset comprises of 70,000 grayscale images, each representing a handwritten digit ranging from 0 to 9 in a resolution of 28x28 pixels. The dataset consists of 60,000 training images and 10,000 testing images.
\paragraph{SVHN}SVHN is a real-world digit recognition dataset obtained from house numbers in Google Street View images. It is similar to MNIST images but the difficulty of recognition for machine learning algorithms is a bit harder.
\paragraph{LSUN}Variations of LSUN datasets are designed for the purpose of scene understanding in large-scale datasets.
\paragraph{Places365}
Places365 is a large-scale scene dataset developed for the purpose of training deep-learning models to understand scenes.
\paragraph{Texture}
The Textures dataset contains images of various textures. It gives a collection of unique images apart from widely available object or scene images. 
\paragraph{TinyImageNet}
TinyImagenet is a smaller version of the larger ImageNet dataset. It consists of 200 classes. The TinyImagenet dataset was created to make research consisting of rich categories computationally feasible with relatively lesser computing infrastructures.

\section{FPR@95 across various OOD datasets}
\begin{figure}[!h]
  \begin{minipage}[t]{\linewidth}    
    \centering    \includegraphics{figures/radar/ood_dataset/legend.pdf}\label{fig:radar_ood_cifar10_base} 
  \end{minipage}

  \begin{minipage}[b]{0.48\linewidth} 
    \centering
    \includegraphics[width=0.8\linewidth]{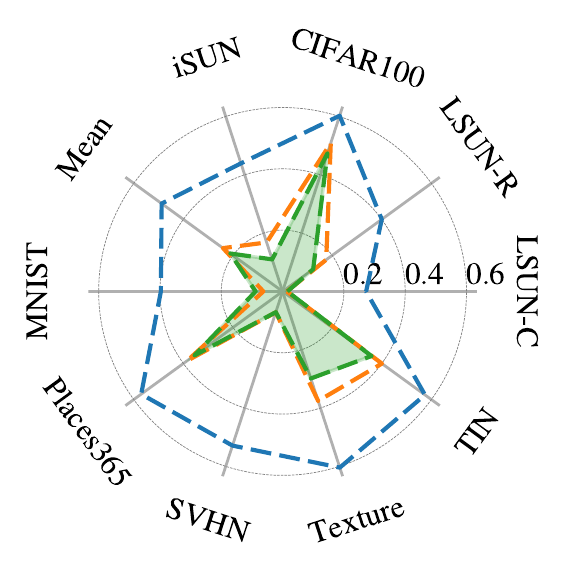}
    \caption{FPR@95 for CIFAR-10 as ID (MSP) (DenseNet)}\label{fig:radar_ood_cifar10_base_densenet}
  \end{minipage}
  \hfill
  \begin{minipage}[b]{0.48\linewidth} 
    \centering
    \includegraphics[width=0.8\linewidth]{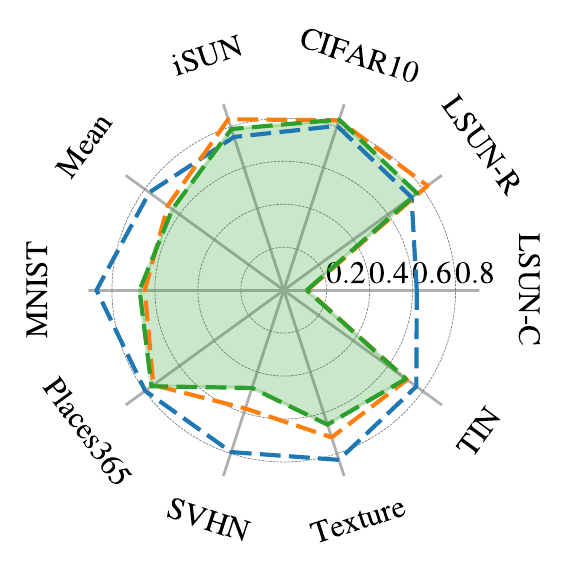}
    \caption{FPR@95 for CIFAR-100 as ID (MSP) (DenseNet)}\label{fig:radar_ood_cifar100_base_densenet}
  \end{minipage}
    
  \begin{minipage}[b]{0.48\linewidth} 
    \centering
    \includegraphics[width=0.8\linewidth]{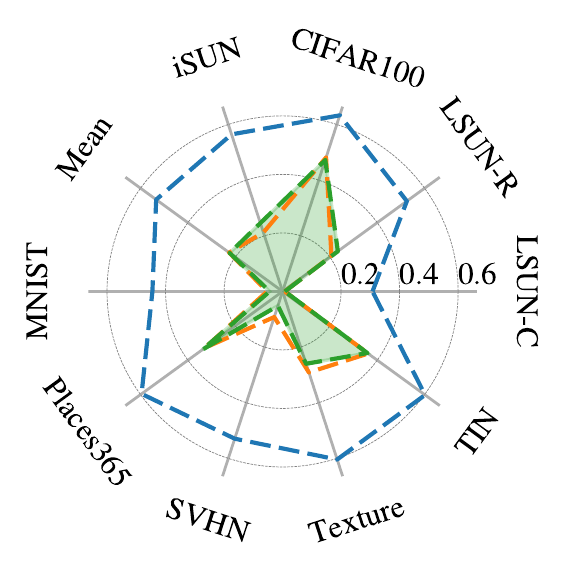}
    \caption{FPR@95 for CIFAR-10 as ID (MSP) (WRN-40-2)}\label{fig:radar_ood_cifar10_base_wrn}
  \end{minipage}
  \hfill
  \begin{minipage}[b]{0.48\linewidth} 
    \centering
    \includegraphics[width=0.8\linewidth]{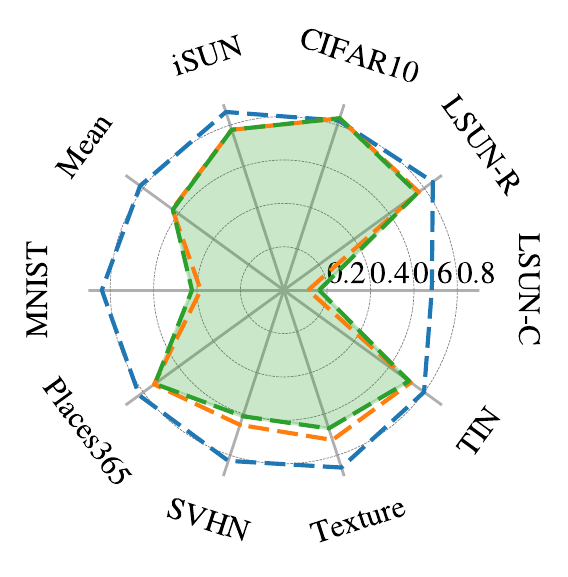}
    \caption{FPR@95 for CIFAR-100 as ID (MSP) (WNR-40-2) }\label{fig:radar_ood_cifar100_base_wrn}
  \end{minipage}
\end{figure}

The FPR@95 metric across various architectures with both CIFAR-10 and CIFAR-100 as ID is shown in the radar plot in Figure \ref{fig:radar_ood_cifar10_base_densenet}, \ref{fig:radar_ood_cifar100_base_densenet}, \ref{fig:radar_ood_cifar10_base_wrn}, and \ref{fig:radar_ood_cifar100_base_wrn}.

Observations show that TrainNorm is as competitive as LogitNorm, if not better, in terms of FPR@95 metric.

\section{Compatibility with various OOD scoring functions}
Table \ref{tab:mean_ood_across_network_and_scoring} shows the comparison of various methods in terms of FPR@95, AUROC, and AUPR metrics. For instance, using a parameter-free EBO scoring function, our method achieves significantly superior performance in comparison with others.

\begin{table}[!h]
\renewcommand{\arraystretch}{3.0}
    \centering
    \caption{Mean OOD metrics obtained through various OOD scoring functions in the form of Baseline / LogitNorm / T2FNorm across various architectures.}
    \label{tab:mean_ood_across_network_and_scoring}
\resizebox{0.98\linewidth}{!}{%
    \begin{tabular}{@{}l|l|ccc|ccc@{}}
        \toprule
         &          &  \multicolumn{3}{c|}{CIFAR-10} & \multicolumn{3}{c}{CIFAR-100} \\
      \cmidrule(lr){2-2} \cmidrule(lr){3-5} \cmidrule(lr){6-8}
         &  Network & FPR@95 $\downarrow$ & AUROC $\uparrow$ & AUPR$\uparrow$ & FPR@95$\downarrow$ & AUROC$\uparrow$ & AUPR$\uparrow$ \\
      \midrule

      \multirow{4}{*}{\begin{sideways}EBO\end{sideways}}
        & ResNet & 37.7/37.0/\textbf{17.9} & 91.5/88.9/\textbf{96.7} & 92.7/89.4/\textbf{96.6} & 77.6/72.6/\textbf{66.6} & 81.0/75.0/\textbf{83.3} & 81.2/75.3/\textbf{82.2}\\ 
        & WRN & 35.4/54.9/\textbf{22.5} & 91.1/85.0/\textbf{95.8} & 92.1/84.1/\textbf{95.7} & 78.0/62.6/\textbf{60.0} & 77.0/81.7/\textbf{84.2} & 78.3/81.9/\textbf{84.4}\\ 
        & DenseNet & 30.3/73.9/\textbf{20.0} & 93.3/86.3/\textbf{96.1} & 93.8/83.2/\textbf{96.1} & 69.2/70.3/\textbf{62.3} & 82.4/75.7/\textbf{83.5} & 83.6/77.0/\textbf{83.9}\\
        \cmidrule(lr){2-8}           
        & Mean & 34.5/55.3/\textbf{20.1} & 92.0/86.7/\textbf{96.2} & 92.9/85.6/\textbf{96.2} & 75.0/68.5/\textbf{63.0} & 80.1/77.5/\textbf{83.6} & 81.1/78.1/\textbf{83.5}\\ 
        \midrule

      \multirow{4}{*}{\begin{sideways}GradNorm\end{sideways}}
        & ResNet & 81.0/25.9/\textbf{18.0} & 59.3/95.0/\textbf{96.6} & 69.9/94.9/\textbf{96.5} & 77.2/71.4/\textbf{65.3} & 71.0/72.2/\textbf{82.0} & 76.3/73.6/\textbf{81.3}\\ 
        & WRN & 70.4/26.6/\textbf{23.4} & 63.0/94.6/\textbf{95.6} & 71.9/94.6/\textbf{95.6} & 87.8/60.4/\textbf{57.2} & 52.3/81.3/\textbf{84.3} & 61.1/81.4/\textbf{84.4}\\ 
        & DenseNet & 48.6/31.6/\textbf{21.3} & 79.5/92.8/\textbf{95.5} & 84.8/92.9/\textbf{95.6} & 76.2/69.9/\textbf{62.6} & 69.2/73.4/\textbf{82.2} & 73.2/74.4/\textbf{82.2}\\
        \cmidrule(lr){2-8}                   
        & Mean & 66.7/28.0/\textbf{20.9} & 67.3/94.2/\textbf{95.9} & 75.6/94.1/\textbf{95.9} & 80.4/67.2/\textbf{61.7} & 64.2/75.6/\textbf{82.8} & 70.2/76.5/\textbf{82.6}\\ 
        \midrule
        
      \multirow{4}{*}{\begin{sideways}Odin\end{sideways}}
        & ResNet & 38.7/19.4/\textbf{17.2} & 87.1/96.4/\textbf{96.9} & 90.7/96.3/\textbf{96.9} & 72.5/69.2/\textbf{66.5} & 82.3/81.6/\textbf{84.1} & 83.3/81.2/\textbf{83.8}\\ 
        & WRN & 43.0/\textbf{20.2}/20.8 & 84.3/\textbf{96.2}/96.1 & 88.4/\textbf{96.4}/96.2 & 73.8/\textbf{60.0}/60.1 & 76.2/84.5/\textbf{84.6} & 79.3/85.1/\textbf{85.3}\\ 
        & DenseNet & 34.7/21.7/\textbf{18.1} & 90.4/95.6/\textbf{96.4} & 92.2/95.7/\textbf{96.6} & 64.9/61.0/\textbf{57.5} & 82.4/83.8/\textbf{85.3} & 84.5/84.8/\textbf{86.5}\\
        \cmidrule(lr){2-8}                   
        & Mean & 38.8/20.4/\textbf{18.7} & 87.3/96.1/\textbf{96.5} & 90.4/96.1/\textbf{96.6} & 70.4/63.4/\textbf{61.4} & 80.3/83.3/\textbf{84.7} & 82.4/83.7/\textbf{85.2}\\ 
        \midrule

      \multirow{4}{*}{\begin{sideways}TempScale\end{sideways}}
        & ResNet & 45.6/22.5/\textbf{19.7} & 91.3/95.9/\textbf{96.5} & 91.7/95.5/\textbf{96.3} & 73.4/66.1/\textbf{61.6} & 81.6/81.9/\textbf{84.6} & 81.4/80.1/\textbf{82.7}\\ 
        & WRN & 45.1/\textbf{22.5}/22.8 & 90.5/\textbf{95.9}/95.8 & 91.0/\textbf{95.7}/95.7 & 70.5/58.2/\textbf{57.9} & 79.2/84.9/\textbf{85.1} & 79.2/84.0/\textbf{84.7}\\ 
        & DenseNet & 39.3/23.2/\textbf{20.1} & 92.6/95.5/\textbf{96.2} & 92.7/95.2/\textbf{96.2} & 66.6/58.7/\textbf{56.5} & 82.4/84.6/\textbf{85.9} & 82.5/83.8/\textbf{85.6}\\
        \cmidrule(lr){2-8}                           
        & Mean & 43.3/22.7/\textbf{20.9} & 91.5/95.8/\textbf{96.2} & 91.8/95.5/\textbf{96.0} & 70.1/61.0/\textbf{58.7} & 81.1/83.8/\textbf{85.2} & 81.0/82.7/\textbf{84.3}\\ 

        \bottomrule
    \end{tabular}
}
\end{table}

\section{Distribution of Norm}
\begin{figure}[h]
  \centering
  \begin{subfigure}{0.34\textwidth}
    \includegraphics{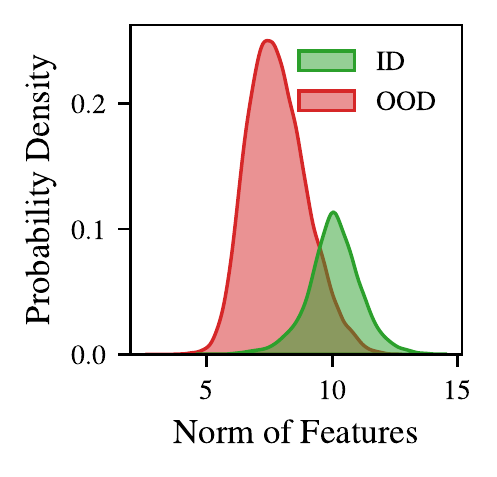}
    \caption{Baseline}
  \end{subfigure}
  \hfill
  \begin{subfigure}{0.32\textwidth}
    \includegraphics{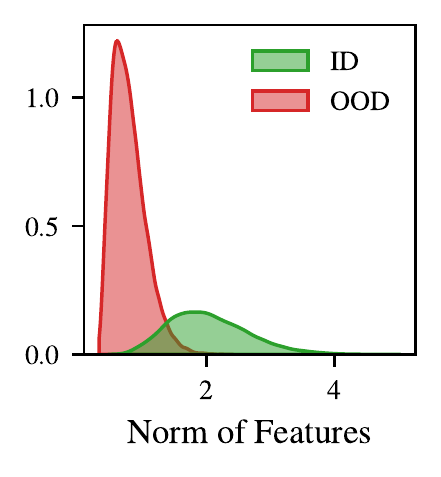}
    \caption{LogitNorm}
  \end{subfigure}
  \hfill
  \begin{subfigure}{0.32\textwidth}
    \includegraphics{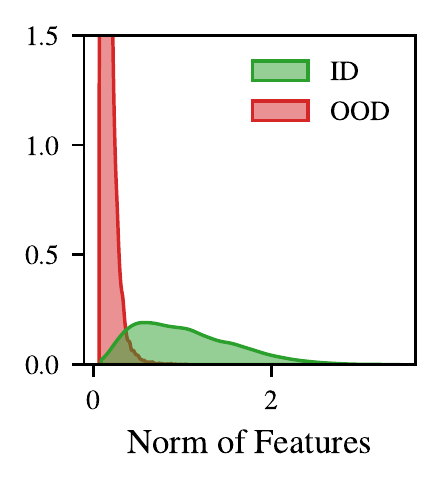}
    \caption{T2FNorm}
  \end{subfigure}
  \caption{Distribution of norm of feature of ResNet-18 model trained with CIFAR-10}\label{fig:fig1_norm}
\end{figure}

\begin{figure}[h]
  \centering
  \begin{subfigure}{0.34\textwidth}
    \includegraphics{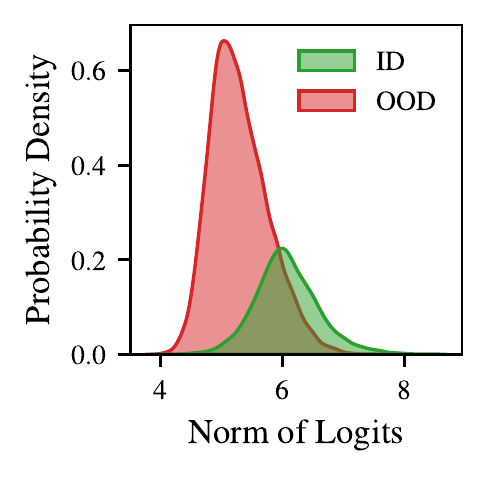}
    \caption{Baseline}
  \end{subfigure}
  \hfill
  \begin{subfigure}{0.32\textwidth}
    \includegraphics{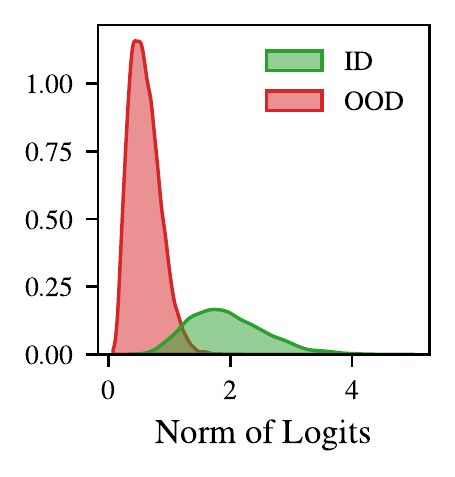}
    \caption{LogitNorm}
  \end{subfigure}
  \hfill
  \begin{subfigure}{0.32\textwidth}
    \includegraphics{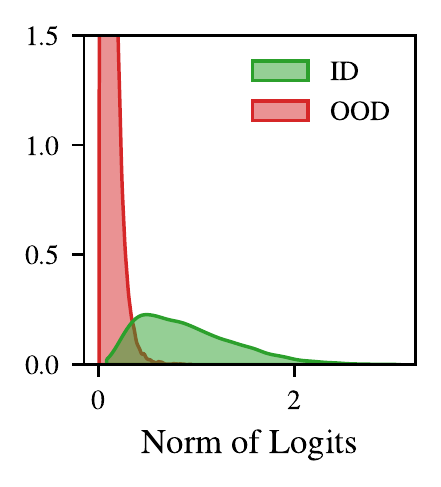}
    \caption{T2FNorm}
  \end{subfigure}
  \caption{Distribution of norm of logit of ResNet-18 model trained with CIFAR-10}\label{fig:fig2_norm}
\end{figure}

The distribution of feature norm for ID (CIFAR-10) and OOD (SVHN) datasets in each of the three methods (Baseline, LogitNorm, T2FNorm) extracted from ResNet-18 architecture are shown in the Figure \ref{fig:fig1_norm}, and \ref{fig:fig2_norm}. In comparison to the baseline, both LogitNorm and T2FNorm have lesser overlap among ID/OOD samples.

\end{document}